%% file: main.tex
\documentclass[acmtog]{acmart}
\acmSubmissionID{258}

\usepackage{booktabs} % For formal tables
\AtBeginDocument{%
  \providecommand\BibTeX{{%
    \normalfont B\kern-0.5em{\scshape i\kern-0.25em b}\kern-0.8em\TeX}}}

\newcommand{\revise}[1]{\textcolor{black}{#1}}

\setcopyright{rightsretained}
\acmJournal{TOG}
\acmYear{2022}\acmVolume{41}\acmNumber{6}\acmArticle{213}\acmMonth{12} \acmDOI{10.1145/3550454.3555462}

\begin{document}
% Title portion
\title{SCULPTOR: Skeleton-Consistent Face Creation Using a Learned Parametric Generator}

\author{Zesong Qiu}
\authornote{Authors contributed equally to the paper.}
\affiliation{%
  \institution{ShanghaiTech University}
  \city{Shanghai}
  \country{China}}
\email{qiuzs@shanghaitech.edu.cn}

\author{Yuwei Li}
\affiliation{%
  \institution{ShanghaiTech University}
  \city{Shanghai}
  \country{China}}
\email{liyw@shanghaitech.edu.cn}
\authornotemark[1]

\author{Dongming He}
\affiliation{
  \institution{Shanghai Ninth People's Hospital}
  \city{Shanghai}
  \country{China}}
\email{1295227946@qq.com}
\authornotemark[1]

\author{Qixuan Zhang}
\affiliation{
  \institution{ShanghaiTech University, China and Deemos Technology Co., Ltd.}
  \city{Shanghai}
  \country{China}}
\email{zhangqx1@shanghaitech.edu.cn}

\author{Longwen Zhang}
\affiliation{
  \institution{ShanghaiTech University, China and Deemos Technology Co., Ltd.}
  \city{Shanghai}
  \country{China}}
\email{zhanglw2@shanghaitech.edu.cn}
\email{zhanglw@deemos.com}

\author{Yinghao Zhang}
\affiliation{%
  \institution{ShanghaiTech University}
  \city{Shanghai}
  \country{China}}
\email{zhangyh5@shanghaitech.edu.cn}

\author{Jingya Wang}
\affiliation{%
  \institution{ShanghaiTech University}
  \city{Shanghai}
  \country{China}}
\email{wangjingya@shanghaitech.edu.cn}

\author{Lan Xu}
\affiliation{%
  \institution{ShanghaiTech University}
  \city{Shanghai}
  \country{China}}
\email{xulan1@shanghaitech.edu.cn}

\author{Xudong Wang}
\authornote{Corresponding author.}
\affiliation{
  \institution{Shanghai Ninth People's Hospital}
  \city{Shanghai}
  \country{China}}
\email{xudongwang70@hotmail.com}
% \authornotemark[2]

\author{Yuyao Zhang}
\affiliation{%
  \institution{ShanghaiTech University}
  \city{Shanghai}
  \country{China}}
\email{zhangyy8@shanghaitech.edu.cn}
\authornotemark[2]

\author{Jingyi Yu}
\affiliation{%
  \institution{ShanghaiTech University}
  \city{Shanghai}
  \country{China}}
\email{yujingyi@shanghaitech.edu.cn}
\authornotemark[2]

\renewcommand\shortauthors{Qiu, et al.}

\begin{abstract} 
Recent years have seen growing interest in 3D human face modeling due to its wide applications in digital human, character generation and animation. Existing approaches overwhelmingly emphasized on modeling the exterior shapes, textures and skin properties of faces, ignoring the inherent correlation between inner skeletal structures and appearance. In this paper, we present SCULPTOR, 3D face creations with \textit{Skeleton Consistency Using a Learned Parametric facial generaTOR}, aiming to facilitate the easy creation of both anatomically correct and visually convincing face models via a hybrid parametric-physical representation. At the core of SCULPTOR is LUCY, the first large-scale shape-skeleton face dataset in collaboration with plastic surgeons. Named after the fossils of one of the oldest known human ancestors, our LUCY dataset contains high-quality Computed Tomography (CT) scans of the complete human head before and after orthognathic surgeries, which are critical for evaluating surgery results. LUCY consists of 144 scans of 72 subjects (31 male and 41 female), where each subject has two CT scans taken pre- and post-orthognathic operations.
Based on our LUCY dataset, we learned a novel skeleton consistent parametric facial generator, SCULPTOR, which can create unique and nuanced facial features that help define a character and at the same time maintain physiological soundness. Our SCULPTOR jointly models the skull, face geometry and face appearance under a unified data-driven framework by separating the depiction of a 3D face into shape blend shape, pose blend shape and facial expression blend shape. SCULPTOR preserves both anatomic correctness and visual realism in facial generation tasks compared with existing methods. Finally, we showcase the robustness and effectiveness of SCULPTOR in various fancy applications unseen before, like archaeological skeletal facial completion, bone-aware character fusion,  skull inference from images, face generation with lipo-Level change and facial animations, etc. 
\end{abstract}




% Computer tomography
\begin{teaserfigure}
    \includegraphics[width=\textwidth]{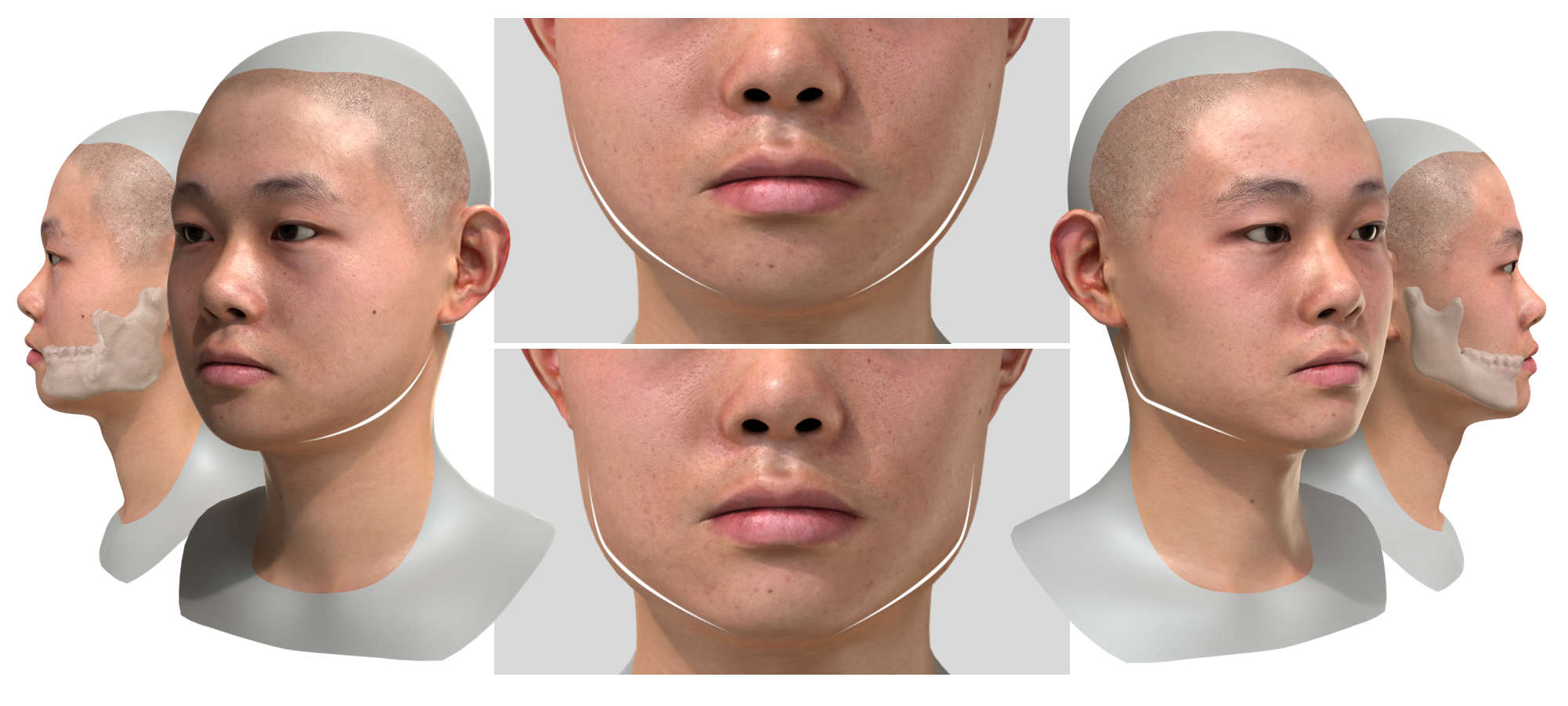}
    \caption{
    We present SCULPTOR, a skeleton-consistent face generator that jointly models the skull, face geometry, and face appearance, allowing for high-quality and characteristic facial detail generation. In this example, we edit the shape and length of the mandible (skeletal structure under the cheek), generating a plump cheek and a square cheek for the same person while keeping the upper face untouched. }
    \label{fig:teaser}
  \end{teaserfigure}

\maketitle

\input{sections/intro}

\input{sections/related}

\input{sections/overview}

\input{sections/data}
\input{sections/method}

\input{sections/application}

\input{sections/experiment}
\input{sections/conclusion}

\begin{acks}
This work was supported by NSFC programs (61976138, 61977047), the National Key Research and Development Program (2018YFB2100500),  STCSM (2015F0203-000-06), SHMEC (2019-01-07-00-01-E00003) and Cultivation of Interdisciplinary Projects (YG2022ZD011).
\end{acks}

% DO NOT INCLUDE ACKNOWLEDGMENTS IN AN ANONYMOUS SUBMISSION TO SIGGRAPH 2019
%\begin{acks}
%
%The authors would like to thank Dr. Maura Turolla of Telecom
%Italia for providing specifications about the application scenario.
%
%The work is supported by the \grantsponsor{GS501100001809}{National
%  Natural Science Foundation of
%  China}{http://dx.doi.org/10.13039/501100001809} under Grant
%No.:~\grantnum{GS501100001809}{61273304\_a}
%and~\grantnum[http://www.nnsf.cn/youngscientists]{GS501100001809}{Young
%  Scientists' Support Program}.
%
%
%\end{acks}

% Bibliography
\bibliographystyle{ACM-Reference-Format}
\bibliography{sample-bibliography}

% % Appendix
% \appendix
% \section{Switching Times}

% In this appendix, we measure the channel switching time of Micaz
% \cite{CROSSBOW} sensor devices.  In our experiments, one mote
% alternatingly switches between Channels~11 and~12. Every time after
% the node switches to a channel, it sends out a packet immediately and
% then changes to a new channel as soon as the transmission is finished.
% We measure the number of packets the test mote can send in 10 seconds,
% denoted as $N_{1}$. In contrast, we also measure the same value of the
% test mote without switching channels, denoted as $N_{2}$. We calculate
% the channel-switching time $s$ as
% \begin{displaymath}%
% s=\frac{10}{N_{1}}-\frac{10}{N_{2}}.
% \end{displaymath}%
% By repeating the experiments 100 times, we get the average
% channel-switching time of Micaz motes: 24.3\,$\mu$s.

\end{document}

%% file: sections/intro.tex
\section{Introduction}

The amazing variety of human faces – far greater than that of most other animals – make each of us unique and easily recognizable~\cite{FacialAttractiveness,RecognizeFace,InvitedUpdate}. From the plump cheeks of the Mona Lisa to strong and finely chiseled chins shared by different versions of Batman, facial traits are the defining characteristics of human characters, real or virtual, physical or digital. To faithfully model human faces, existing approaches have overwhelmingly emphasized modeling the exterior shapes, textures and skin properties of faces. Physically-based techniques based on photometric or multi-view 3D scanning can now recover ultra-fine geometry at pore-level. Yet, they generally require using bulky and expensive apparatus~\cite{MultiviewFaceCap} and have been limited to celebrities for feature film productions. By far, 3D scanned models are still of much fewer varieties than real ones. 

To enrich the diversity of facial models, tremendous efforts have been focused on developing easy-to-use face generators, ranging from earlier parametric models such as 3DMM~\cite{blanz1999morphable_3dmm} to the latest data-driven model such as FLAME~\cite{flame} and DECA~\cite{feng2021learning_deca}. Despite their effectiveness, few techniques employ anatomic facial bone structures in the model generation process. Physiologically, facial appearance is biomechanically related to skeletal structures: bones grow under pulling force and absorb under pressure; consequently, strong muscles are often matched with thick bones, inducing characteristic contours and features. Putting aside biomechanical rationality can produce absurd results that are far less convincing than an anatomically \revise{consistent} one. Therefore the variety of faces has boundaries and should conform to the anatomical rules. For face generators, it is hence crucial to provide easy-to-use controls over the facial skeleton structures while abiding physiological soundness.

%Therefore, introducing real bone information to limit the variation range can avoid illogical virtual face modeling.

%In reality, Dr. XXX. The inverse of inferring facial structures from images is more challenging: without skeletal priors, the problem is ill-posed as the same exterior facial geometry may correspond to drastically different bone structures. Therefore, an anatomical face model that correlates appearance with skeletal structures is in great demand for both generation and inference tasks.   

%inference techniques are  In addition, the modeling of interior skeletal structures, which is critical for accurate movement and expressions, is largely missing. facial skeletons play a determining role in the final appearance of human faces. 
% \todo{Dr. He, please fill in the comments.}
% XXX: JAW BONE怎么怎么影响shape，颧骨怎么XXX， 请何医生fill in. 
% 

Facial skeletons further serve as a key invariant to appearance: the body weight and muscle composition as well as skin textures and elasticity of the same character may change over different periods of time but its skeleton geometry remains largely unchanged. Given the renowned actor Christian Bale as an example whose presence in the feature films from "The Mechanic" to "The Batman" and to "Vice" went through drastic body weight changes leading to dramatic facial appearance variations, his facial skeletons can serve as a constraint to help disentangle shapes, bones, and appearance as well as enable interpolation or extrapolation of respective attributes for future auditions.   

% paragraph 1
% 1. Facial assets. usefulness in production, gaming, and metaverse. Existing pipeline: 1) capture using dome systems or lightstage, expensive, only for celebrity; 2) artist, labor intense; require elaborate skills. Still lacks diversity: beauty should be diverse and not restricted to stereotypes; 

% paragraph 2
In this paper, we present SCULPTOR, 3D face creations with \textit{Skeleton Consistency Using a Learned Parametric facial generaTOR} that we derive from comprehensive anatomical studies. SCULPTOR aims to facilitate the easy creation of both \revise{being anatomically consistent} and visually convincing face models via a hybrid parametric-physical representation. 
% A number of parametric face models have been proposed in the past decade. 3DMM. DECA. 
% 2. Existing facial generation tools do not consider anatomical correctness. Only appearance (geometry), can easily fall into uncanny valley. Facial characters such as cheek, chin, jaw, diverse but realistic. Explicit: 3DMM limited shape and realism. Implicit: Deca does not provide explicit controls over bone structures

% qiu
% paragraph 3
At the core of SCULPTOR is LUCY, the first large-scale shape-skeleton face dataset in collaboration with plastic surgeons. Named after the fossils of one of the oldest known human ancestors, our LUCY dataset contains high-quality Computed Tomography (CT) scans of the complete human head before and after orthognathic surgeries, which are critical for evaluating surgery results. CT, as a 3d medical imaging technique, is widely used in orthodontics for diagnosis, treatment planning, mock surgery and post-treatment assessment~\cite{agrawal2013cbct}. 
LUCY consists of 144 scans of 72 subjects (31 male and 41 female), where each subject has two CT scans taken pre- and post-orthognathic operations. 
Specifically, we obtain accurate 3D surgical landmarks and segmentation annotations of the internal mandible and maxilla skeleton and on the external facial geometry, both labeled by experienced plastic surgeons. 
To correlate inner skeletal structures with appearance, we also acquire the exterior 3D facial geometry along with texture maps. Specifically, we employ {3dMD~\cite{3dmd}, a structured light-based multi-view RGBD scanning system, to recover initial 3D facial geometry with skin textures before and after operations.}
%
% 3. We present a novel skeleton driven facial asset generation framework. At the core, a novel facial skeleton dataset. Details: from plastic surgeons. Correct key bone structures that affect most the final appearance. how many subjects, IBR (anonymized), 3D scan both external geometry and CT bone structure; By analyzing how bones affect appearance, construct a novel skeleton-driven facial asset generator. 

% template, registration, parameter learning, and diversity!
{
Next, we utilize a general neutral head mesh that consists of mandible, maxilla and outer surface mesh as the template geometry and set out to learn the parametric SCULPTOR model.
SCULPTOR aims to separate the depiction of a 3D face into shape blend shape, pose blend shape and facial expression blend shape. Therefore we conduct mesh registration and model learning analogous to techniques used on exterior faces, hands, and even full body shapes ~\cite{smpl,mano,flame} to train SCULPTOR parameters including skinning weight and various blend shapes. 
Validations on the ground truth pre- and post-orthognathic surgery data further demonstrate that SCULPTOR is reliable and accurate. In particular, compared with prior art ~\cite{flame,blanz1999morphable_3dmm}, SCULPTOR can create unique facial features that help define a character while maintaining physiological soundness. 
% 因为是在真实的医疗数据上训练的，所以这个generator捕捉到了真实人脸diversity distribution/variation.
}

% qiu
% paragraph 4
% 4. Enable a new class of applications. Bone-stitcher for new facial asset creation. Details algorithm. Analyze the same celebrity over BMI changes, predict BMI, enable for facial animations. Detailed algorithm. Combine with StyleSDF to create FVV portrait that follow the desired bone structure. Detailed algorithm. 

{
The skeleton-consistent nature of SCULPTOR benefits a variety of applications including: 
% Archaeological Skeletal Facial Completion
(1) Facial geometry estimation from incomplete facial bone structures that can be obtained as near as from a partial CT scan and as far as through archaeological explorations. 
% Character fusion
(2) Augmenting existing facial assets by adjusting exterior face geometry or even fusing facial appearance from different characters while enforcing to \revise{be anatomically consistent}. 
% as others, inference using optimization/regression
(3) Face/skull inference from scanned face models or even images, by employing the differentiable network layer analogous to ~\cite{flame,feng2021learning_deca}.  
% Body Mass Index(BMI) change - 对body fat进行建模，而不是紧贴骨头的与fat无关的部分。
(4) Supporting physically correct facial animations of drastic head/face movements as well as under external forces by first inferring and then imposing inner skeletal structures as constraints. 
}

% qiu
% paragraph 5
% 5. Comparison to SOTA. more details on accuracy and visual realism, physical simulation. Diversify facial assets with anatomic correctness. 
%\lyw{
%We conduct qualitative and quantitative experiments on our model, and compare it with %state-of-the-art models on generalization ability. Our model outperforms others on both visual realism and evaluation results. 
%}

To summarize, our main technical contributions include:
\begin{itemize} 
	\setlength\itemsep{0em} 
	% dataset
	\item {We present LUCY, \revise{a} comprehensive shape-skeleton correlated face dataset from pre- and post-surgery CT imaging and 3D scans. LUCY contains rich annotations on surgical landmarks and semantic segmentation labels and will be disseminated to the community after de-identification and anonymization.}
 % \item {We present LUCY, a comprehensive \revise{pre- and post-surgery CT imaging and 3D scans dataset, enabling} shape-skeleton correlated face dataset. LUCY contains rich annotations on surgical landmarks and semantic segmentation labels and will be disseminated to the community after de-identification and anonymization.}
	
	% model: inner+outer, diversity
	\item {We derive a skeleton consistent face generator model SCULPTOR from LUCY that jointly models the skull, face geometry, and face appearance under a unified data-driven framework. Compared with the SOTA, SCULPTOR preserves both anatomic correctness and visual realism in facial generation tasks. }
	
	% applications
	\item {We apply SCULPTOR to aforementioned applications and demonstrate its robustness and effectiveness. In particular, SCULPTOR helps to enrich currently scarce 3D face data with physical correctness. }

\end{itemize}

%% file: sections/related.tex
\section{Related Work}

%\cite{herlihy:methodology}

In this section, we review contemporary related studies on Data acquisition for Parametric models, 3D Parametric Face Models and Anatomically-Constrained Parametric Face Models.

% In this section, we survey closely related works and discuss the relationship with the proposed work. 

% we survey closely related works and discuss the relationship with the proposed work. 
% XXX: 是否考虑写一段医学的；可以加分
%%%%%%%%%%%%%%%%%%%%%%%%%%%%%%%%%%%%%%%%%%%%%%%%%%%%%%%%%%%%%

%\noindent\textbf{Parametric Models} 
\noindent\textbf{Data acquisition for Parametric models. } 
% \noindent\textbf{Parametric Models} 
To synthesize the surface of faces with different shape, pose and appearance, parametric face models \cite{flame,Local-Deform-Face,high_fidelity3d,Complete-Morphable} estimate low-dimensional parametric space to approach 3D geometry. It assumes that human body shape geometries lie on a manifold, which can provide prior statistical knowledge that helps to solve ill-posed vision problems. The primary ingredient of the parametric model is a representative set of 3D shapes, usually coupled with corresponding appearance data acquired from the real world ~\cite{Morphable}.
Laser scanners, time-of-flight sensors, multi-view photogrammetry and structured light systems are commonly used for 3D face data acquisition. Subsequently, geometric, photometric and hybrid methods are applied for capturing facial shape information from the data scanning. The facial appearance can also be constructed via back-projecting methods using reconstructed facial surface mesh as prior guidance. 

Recently, tremendous technical improvement has been made for the acquisition of facial performance and morphology, especially on detailed skin micro-structure ~\cite{Skin-Microstructure,Appearance} , hair ~\cite{Hair-Modeling}, eyes \cite{Eye-Capture}, eyelids ~\cite{Eyelids,Eyelids-Tracking}, beards ~\cite{Beard}, lips ~\cite{Lips}, teeth ~\cite{Teeth} and tongue \cite{Tongue} . In addition, medium-scale details (wrinkles) are captured from monocular input in real-time ~\cite{Facial-Capture,LiveCap}. Anatomical constraints have proven useful in estimating the rigid transformation of the skull (rigid stabilization)
~\cite{disney_stable,PhaceICHIM,Local-Deform-Face,CVPR_ProbabilisticJointFaceSkull} and extracting detailed flesh deformations ~\cite{Local-Deform-Face}. However, limited by the light field data capture systems, real relationships between facial skeleton, shape and appearance are not well explored. 

\begin{figure*}[t]
    \centering
    \includegraphics[width=1\linewidth]{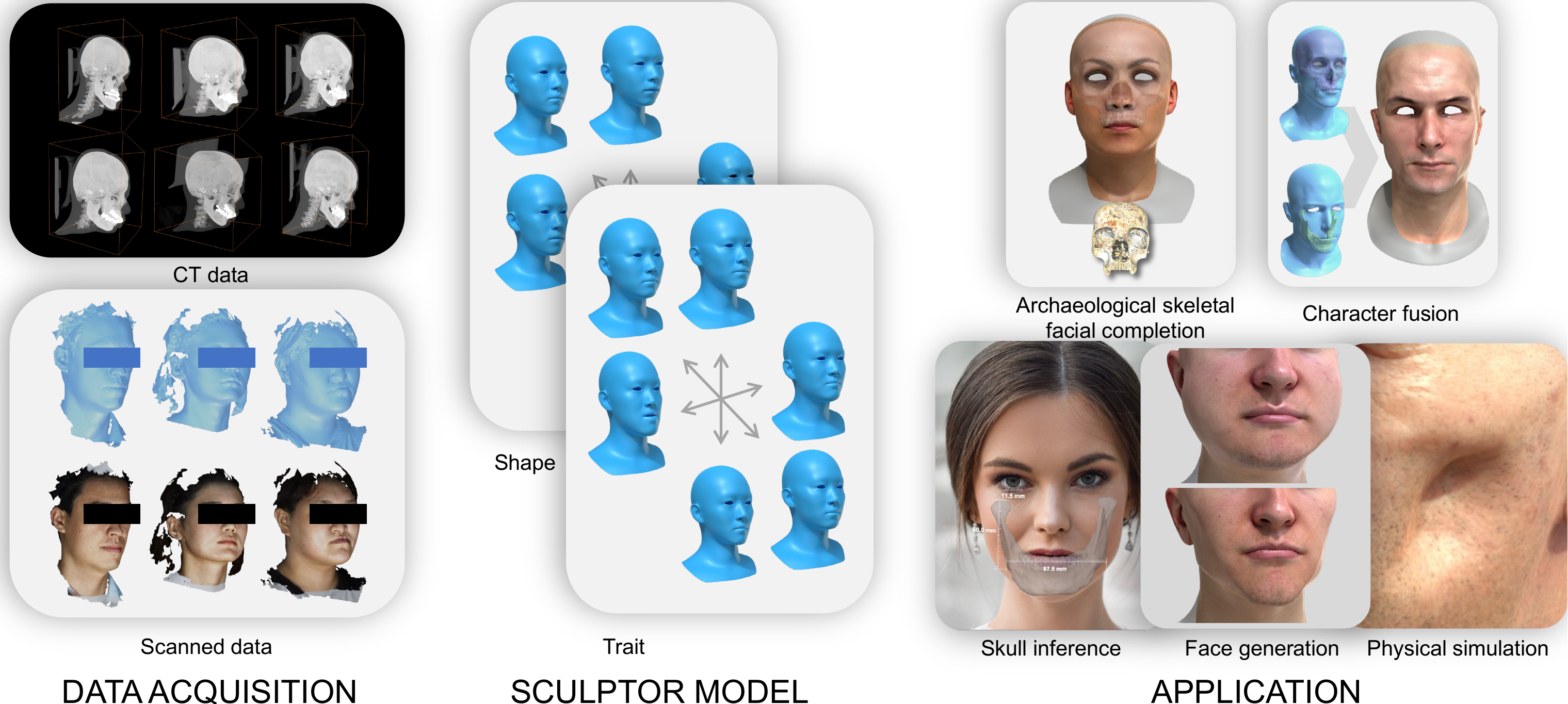}
    \caption{Overview of building SCULPTOR, which includes mesh registration to all the interior and exterior face features as well the photometric appearances in the LUCY dataset; and the parametric model training process over the dataset.  SCULPTOR learns a skeleton-consistent face model that can generate much more diverse faces while following anatomic principles. The rendered skull in Archaeological skeletal facial completion image from Open Virtual World (Sketchfab). The image in Skull inference from T. Bellis (Unsplash).}
    \label{fig:Overview}
\end{figure*}

The most common imaging techniques for acquiring facial anatomical structures are computed tomography (CT) and magnetic resonance imaging (MRI). CT presents superior clear skeleton and face contour information but is constrained by its high radiation exposure with only clinical allowance. MRI scan is a safe solution that can provide good contrast in human body soft tissue (such as muscle, fat, tender and neuron) ~\cite{Contrast}. However, the complex structure of human soft tissue and unclear skeletal structure in MRI make it difficult to extract anatomical structures related to facial appearance. Therefore, most existing anatomical constraints parametric face models are based on artificial skulls ~\cite{disney_stable} or public CT head dataset \revise{\cite{ct400}}. \revise{In \cite{ct400}, it provides CT head scans 
from 400 patients. However, the dataset is not able to model how the face changes when part of the skull is modified.}

\noindent\textbf{Parametric Face Models. }
\revise{\cite{skullparametric,multilinear_skullface} have built a statistical skull-face shape model with 114 CT scans, enabling a mapping function from skull to face. However, it has not equipped the face model with pose variation space. }
Most of the current \revise{pose-enabled} parametric face models focus on the face's outer surface, shape and texture information \cite{flame,Local-Deform-Face,high_fidelity3d,Complete-Morphable}. For example, FLAME ~\cite{flame} learned a comprehensive facial shape and expression model that captures a wide range of outer facial shapes. For generating more realistic geometries, some exaggerated expressions such as shouting and laughing, FLAME considered an abstract jaw articulation as an axis of mandibular changes. However, without training with precise jaw articulation location, expressions with large jaw motions generated by FLAME still lack realism. 
~\cite{high_fidelity3d} created a differentiable parametric human head model to recover high-resolution realistic facial geometries and texture of the whole head's outer surface. Without specific variations in facial expression and jaw motion, the model tended to generate facial geometry with only gentle poses and expressions. These models are typically global, meaning that the entire face is parameterized holistically.
Local or region-based parametric face models have also been proposed, which offer more flexibility at the cost of being less constrained to realistic human face shapes. ~\cite{Multilinear-Wavelets} used many localized multi-linear models to reconstruct faces from noisy or occluded point cloud data. ~\cite{Sparse-Local} extracted sparse localized deformation components from an animated mesh sequence, also with the goal of intuitive editing as well as statistical processing of the face. ~\cite{Local-Deform-Face} proposed a local 3D face model that parameterizes the face into many overlapping patches and explicitly encodes the local deformation of each patch rather than local positions, which allows monocular face reconstruction and single-view direct editing with unprecedented fidelity. ~\cite{feng2021learning_deca} introduced an animatable detailed 3D face model called DECA that disentangles person-specific details from expression-dependent wrinkles allowing the proposed model to synthesize realistic person-specific wrinkles by controlling expression parameters while keeping person-specific details unchanged.

%%%%%%%%%%%%%%%%%%%%%%%%%%%%%%%%%%%%%%%%%%%%%%%%%%%%%%%%%%%%%%%%%%%%%%
%\noindent\textbf{Parametric Face Models with Anatomical Structure Constraints.}  
The internal skeletal structure of the face highly affects facial appearance. Therefore, physical or anatomical constraints are essential for modeling face pose and shape with large variation space. 
In ~\cite{PCA-Facial-Shape}, the authors created two independent statistical shape models for the skull and face, respectively. The skull models were built from a large dataset of CT head scans, and a linear mapping was found between the two models, following a common pipeline from automatic forensic facial reconstruction ~\cite{Craniofacial-Recon}. 
In ~\cite{CVPR_ProbabilisticJointFaceSkull}, the authors proposed to solve the problem of aligning an independent face model to a skull model with stochastic optimization. A probabilistic joint face-skull model is then constructed, which is able to infer a distribution of plausible face shapes given a skull shape. 
Not only the facial skeleton but other kinds of attributes also play an important role in generating a more realistic face shape and appearance. In several approaches, researchers suggest changing the facial attributes in a final step to adjust soft tissue variation due to age and Body Mass Index (BMI) differences as a complementary approach ~\cite{disney_skull_gen}. For example, by integrating a template skull model together with underlying facial muscle, Phace ~\cite{PhaceICHIM} produced physically-correct facial animation performance. 
~\cite{ict} added teeth, gums, eyeballs, eye blending, lacrimal fluid, eye occlusion, and eyelashes into the face model to the ensure anatomical correctness of the generated digitized face assets.
In face capture tasks, ~\cite{Local-Deform-Face} used the underlying bone structure to anatomically constrain the local skin thickness, simultaneously solving for the skin surface and the skull position for every video frame, yielding a rigidly stabilized performance. Although with limitations, the anatomical structure is indicated to be an essential component in leading a physically correct facial animation performance. 

Inspired by a recently proposed parametric hand bone model from an MRI dataset ~\cite{li21piano}, which achieved inner hand kinetic structure in a data-driven manner. We consider the use of medical imaging techniques to build a novel face generator that jointly models the skull, face geometry, and facial appearance to better facilitate our modification of facial appearance by anatomical structure and generate more realistic and diverse faces.

\begin{table*}[htb]
    \centering
    \caption{SCULPTOR vs. existing face  models.}
    \begin{tabular}{c|c|c|c|c|c|c|c|c|c}
    \hline
        Model  & Parametric & Skull & Face & Anatomically Consistent & Shape & Pose & Expression & Appearance& Trait\\ 
        \hline 
        % \hline

        \cite{CVPR_ProbabilisticJointFaceSkull} &  {\checkmark} &  {\checkmark} &  {\checkmark} &  {\checkmark} &  {\checkmark} &  {$\times$} &  {$\times$} &
         {$\times$}&
         {$\times$}\\
        % \hline
        \cite{disney_skull_gen} &  {\checkmark} &  {\checkmark} &  {\checkmark} &  {$\times$} &  {\checkmark} &  {$\times$} &  {$\times$}&  {$\times$}&
         {$\times$}\\
        % \hline
        \cite{PhaceICHIM}  &  {$\times$} & {\checkmark}  &  {\checkmark} &  {$\times$} & {\checkmark} & {\checkmark} &  {\checkmark}&
         {\checkmark}&
         {$\times$}\\
        % \hline
        \cite{ict} &  {$\times$} & {$\times$}  &  {\checkmark} &  {$\times$} & {\checkmark} & {\checkmark} &  {\checkmark}&  {\checkmark}  &
         {$\times$}\\
        % \hline
        \cite{flame} &  {\checkmark} &  {$\times$} & {\checkmark} &  {$\times$} & {\checkmark} & {\checkmark} &  {\checkmark}&  {$\times$}&
         {$\times$} \\
        % \hline
        SCULPTOR (Ours) &  {\checkmark} &  {\checkmark} & {\checkmark} & {\checkmark} & {\checkmark} & {\checkmark} &  {\checkmark} &  {\checkmark}
        &  {\checkmark}
        \\ \hline
    \end{tabular}
    \label{tab:model_compare}
    % \vspace{-5mm}
\end{table*}

%% file: sections/overview.tex
\section{Overview}
\begin{figure*}[htb]
    \centering
    \includegraphics[width=\linewidth]{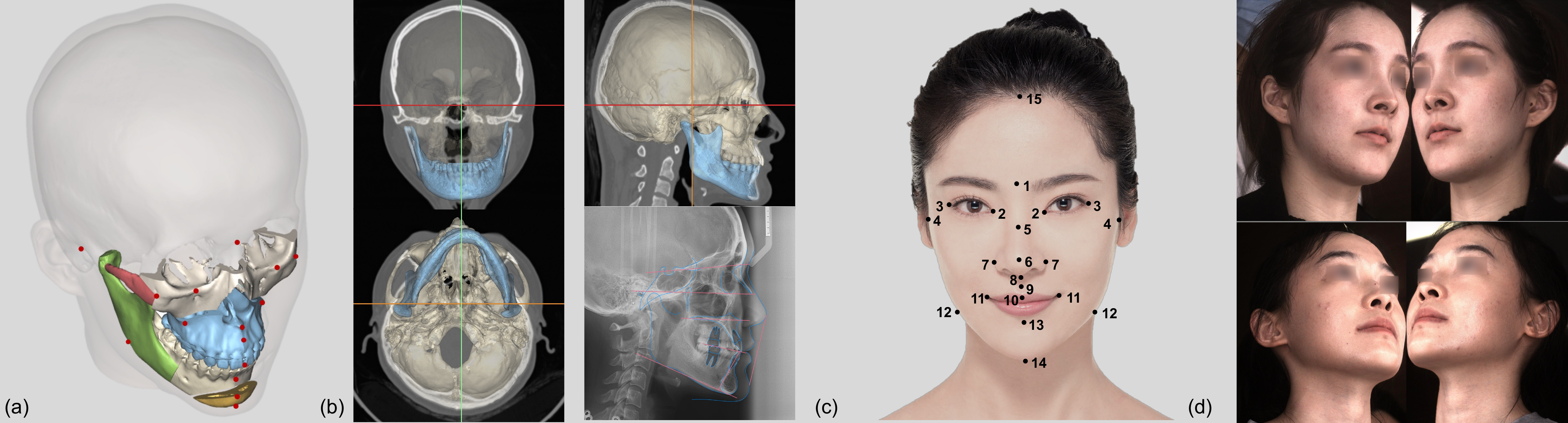}
    \caption{(a) 29 skeletal landmarks labeled on CT scan. Skeletal structures modified during Orthognathic surgery are marked with different colors. (b) Orthogonal slices of CT scan, indicating separated maxilla (white) and mandible (blue). (c) 15 facial landmark positions labeled on face appearance scans. (d) Facial appearance scan using the 3dMDface system in preparation for orthognathic surgery.}
    % \todo{Qiu add figure index and before and after surgery label into figure}
    \label{fig:Data}
\end{figure*}

In this work, we propose SCULPTOR, a novel face generator that jointly models the skull, face geometry, and face appearance under a unified data-driven framework. 

The overview of our method is shown in Fig.~\ref{fig:Overview}. 
SCULPTOR is developed from a large-scale shape-skeleton face dataset LUCY that contains high-quality Computed Tomography (CT) scans of the complete human head pre- and post-orthognathic surgeries. Accurate 3D surgical landmarks and segmentation annotations on the internal mandible and maxilla skeleton and the external facial geometry are labeled by experienced plastic surgeons. Additionally, exterior 3D facial geometry along with texture maps are scanned using 3dMD. 

We then utilize a general neutral head mesh that consists of mandible, maxilla and outer surface mesh as the template geometry to learn the statistical skeleton-driven facial model from LUCY. 
We conduct mesh registration to all the interior and exterior face features and photometric appearances in the dataset. We train SCULPTOR parameters including skinning weight and various blend shapes via model learning techniques used on exterior faces, hands, and even full body shapes ~\cite{flame,mano,smpl,li2022nimble}. 
SCULPTOR learns a skeleton-consistent face model that can generate a much more diverse facial appearance while following anatomic principles.

The rest of the paper is organized as follows: we first introduce our data collection and annotation process in Section~\ref{sec:Data}. Then, we present the model formulation in Section~\ref{sec:model_formulation}, followed by 
model registration and parameter learning 
% physically based registration and multi-stage parameter learning 
on shape, trait, pose and appearance in Section~\ref{sec:parameter_learning}. %Based on the skull consistent face template, we attach appearance to get photo-realistic rendering effect
The procedure of the skeleton-driven face generation and editing
% various faces as well as editing faces 
is shown in Section~\ref{sec:rendering}. 
In Section~\ref{sec:application}, we demonstrate the effectiveness of SCULPTOR on a variety of applications.

%% file: sections/data.tex
% \section{Data Acquisition}
\section{Building LUCY}
\label{sec:Data}

% \begin{figure}[t]
%     \centering
%     \includegraphics[width=\linewidth]{NavieFig/he4.png}
%     \caption{\qzs{Morphological characteristics of face}}
%     % 数字的必要性
%     \label{fig:necessarylmk}
% \end{figure}

%%%%%%% Sub-Section 1 %%%%%%%%%%%%
\subsection{Data Acquisition and Original Usage}
%%%%%%%%%%%%%%%%%%%%%%%%%%%%%%%%%%%%%%%%%%%%%%%%%%%%%%%%
\noindent\textbf {Data original usage background.}
We actively collaborate with orthognathic surgeons to collect the real-world data that shows the skeleton consistent variation of the facial outer surface.
% 
% Orthognathic surgery is a plastic surgery that optimizes facial proportions to treat functional problems caused by bite discrepancies or facial imbalances. It repositions the maxilla, mandible and zygoma through the procedures in Fig.~\ref{fig:procedure}.
% 
The medical images collected during the surgery planning and recovery period clearly depict the influence of facial skeleton changes on the facial appearance, especially on the face's outer surface. To achieve the best orthognathic surgery performance, each patient underwent two CT scannings, pre and post-surgery in Fig.~\ref{fig:Data}(b), as well as multi-view facial scans captured by the imaging system in the hospital as shown in Fig.~\ref{fig:Data}(d).
This routine examination captures the patient skeleton structure and facial appearance features. 

\noindent\textbf {Data acquisition parameter.}
To avoid unnecessary radiation exposure, we retrospectively adopt CT scans from archived medical records at the Department of Oral \& Craniomaxillofacial Surgery, Shanghai Ninth People's Hospital, Shanghai Jiao Tong University School of Medicine. 
In all, a total of 72 individual subject head CT image pairs (pre and post-surgery), as well as the multi-view face appearance scans are collected. The 3D maxillofacial CT imaging was performed using a spiral CT scanner (Light speed 16; GE, Gloucestershire, UK), with image spatial resolution $0.48 \times 0.48 \times 1~mm^3$. These real-world data help us to build a more realistic parametric model.

%%%%%%% Sub-Section 2 %%%%%%%%%%%%%%%%%%%%%%%%%%%%%%%%%%%%%%%%%%%%%%%%%%%%%%%%%%%%%%%%%%%%%
\subsection{Data Labeling}
%%%%%%%%%%%%%%%%%%%%%%%%%%%%%%%%%%%%%%%%%%%%%%%%%%%%%%%%%%%%%%%%%%%%%%%%%%%%%%%%%%%%%%%%%%%%%%%%%%%%%%%%%%%%%%%%%%
The CT volume is a regular volumetric grid of scalar values representing tissue mass density. To analyze anatomical structure, bone and joint location must be annotated. 
To acquire anatomical structure from raw CT data, specialists segment the skull and face from CT volume with thresholding method and morphological operations.
%Experienced surgeons separate skull and soft tissues from the raw CT images using a thresholding method.
Besides, tissues around condyle structures are carefully annotated to break the connection between mandible and maxilla, finally acquiring separated mandible, maxilla volume and the facial outer surface. 
The CT volume and multi-view face scan for orthognathic surgery are both in neutral pose. We thus merge the multi-view scan face reconstruction with the CT facial geometry to add more facial details that are smoothed out during CT scanning. Specially, we apply ICP~\cite{icp} to align multi-view scans to the facial soft tissues captured in CT. 
Thus, we obtained the anatomically \revise{consistent}, detailed facial scan and skull. 
% In order to get more facial details, the CT and light field system capture the same action series. %医院提供的光场数据和CT是相同的动作，做这一步的原因是因为本身CT对外表面的细节capture的不够，用有重建的，可以获得更多脸部细节%The CT volume is a regular volumetric grid of scalar values representing tissue mass density.
%To acquire editable anatomical structure from raw CT data (DICOM), specialists reconstruct head with surface rendering technology. Then experienced surgeons segment skeleton and soft tissue with thresholding method. Finally separated mandible, maxilla and the facial outer surface are extracted and stored as geometry mesh.
Then, biological meaningful landmarks are manually annotated on the mandible, maxilla and face surface for presurgical planning; in our work, 29 skeleton and 15 face surface landmarks are selected as semantic landmarks for model registration. Fig.~\ref{fig:Data}(a)(c) show the skeletal and facial landmark positions, respectively.

%% file: sections/method.tex
\section{SCULPTOR Model}
\label{sec:methodology}
%%%%%%%%%%%%%%%%%%%%%%%%%%%%%%%%%%%%%%%%%%%%%%%
\begin{figure*}[t]
    \centering
    \includegraphics[width=1\linewidth]{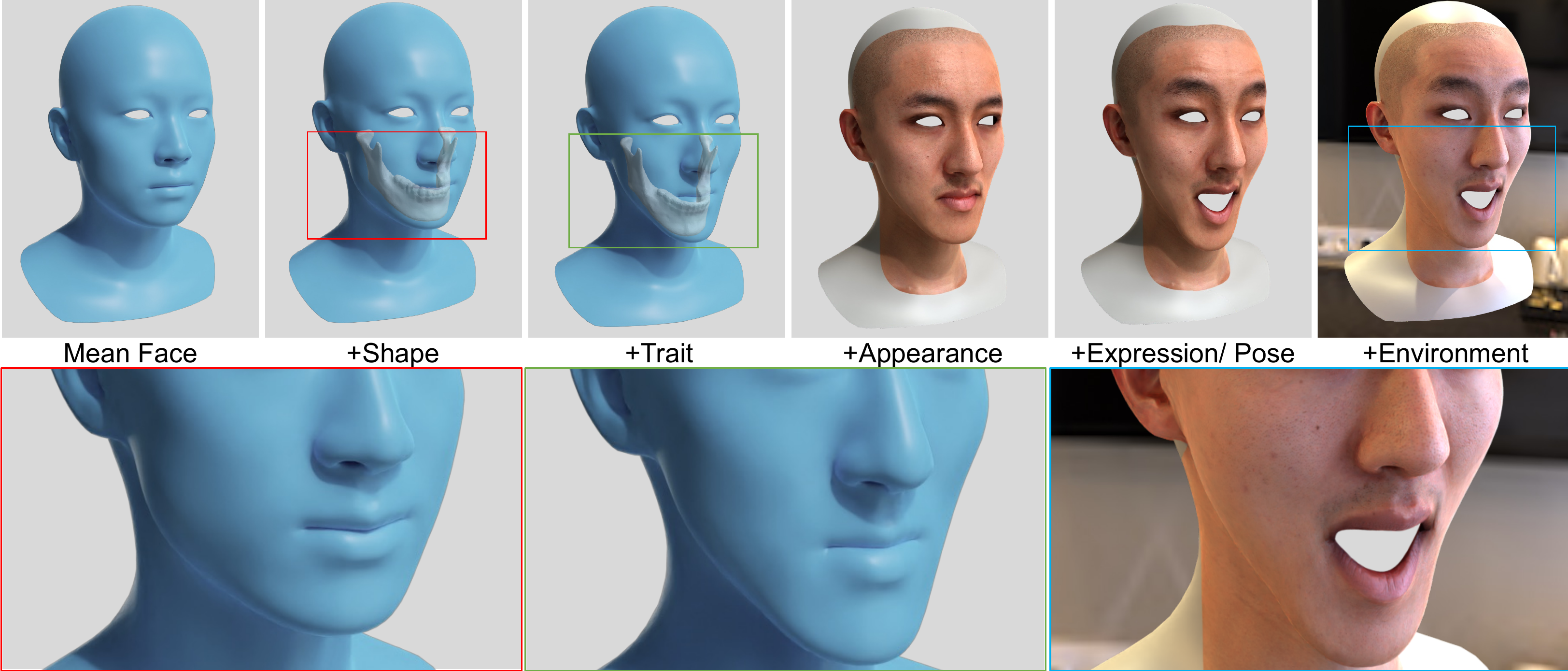}
    \caption{Our realistic face generation pipeline with trait effect. Starting with SCULPTOR full template, we randomly generate and procedurally add shape, trait, appearance and expression/pose effects on the neutral template, rendering the 3D face with environment maps. }
    \label{fig:pipeline}
\end{figure*}
%%%% Global motivation for model construction Yuyao
The interior skeletal structures determine the exterior face shape, geometry, and skin properties. Inspired by this, we intend to build a skeleton-driven parametric face model that provides expressive control in facial skeletons to achieve faithful exterior geometry and a realistic appearance. To achieve this goal, we jointly model the internal face skeletal structure with face exterior shape and geometry. Besides, we add a skeleton modification module trained on the pre- and post-orthognathic surgery medical image data, which is called the 
characteristic generator. The generator significantly enhances the model's ability to represent and control skeletal structures with its influence on face shape.
% and geometry.

%%%%% Model Formulation%%%%%%%%%%%%%%%%%%%
\subsection{Model Formulation}
\label{sec:model_formulation}
%%%%%%%%%%%%%%%%%%%%%%%%%%%%%%%%%%%%%%%%%%

We define our model using the general formulation as follows:
\begin{equation}
    \mathcal{M}(\theta, \beta, \gamma, \phi, \alpha) = \{\mathcal{G}(\theta, \beta, \gamma, \phi),  \mathcal{A}(\alpha)\},
    \label{eqn:full_formulation}
\end{equation}
where $\mathcal{G}$ denotes the geometry for both skeleton and face, and $\mathcal{A}$ models the face appearance. $\theta$, $\beta$, $\gamma$, $\phi$, $\alpha$ are parameters controlling pose, shape, trait, expression and appearance, respectively.

In our model, we propose a parametric face model with more accurate appearance variation with the inner skeleton structure by extending the human face model FLAME~\cite{flame} to a physically precise parametric human skeleton model with the canonical pose. Additionally, to further enhance the ability to control human face appearance using the skeleton, we introduce a novel parameter $\gamma$, which is called the "trait component", to represent a physiologically reasonable variation space for the internal skeleton of the face, as well as face shape variations that caused by the change of skeleton. The template is formulated as $\mathcal{G}$: 
\begin{equation}
    \mathcal{G}({\beta}, {\gamma},{\theta}, {\phi}) = LBS(\mathcal{W}, J_p({\beta},{\gamma}), \mathbf{T}_p({\beta}, {\gamma}, {\theta}, {\phi})),
    \label{eqn:formulation}
\end{equation}
%%%%%%
% \todo{To be modified by Qiu}
where $LBS(\cdot)$ demotes the Linear Blend Skinning (LBS) function; 
$\mathcal{W}$ is the learned skinning weight; 
${\beta}$, ${\gamma}$, ${\theta}$ and ${\phi}$ represent the PCA coefficient vector of the shape, trait, pose and expression space, respectively; 
$\mathbf{T}_p$ represents the person-specific head mesh with corresponding 
% shape, pose, trait and expression 
variation over the general template $\overline{\mathbf{T}}$. 

$J_p$ represents the anatomical joint location for jaws, defined as $J_p({\beta},{\gamma})$=$\mathcal{J}(\overline{\mathbf{T}}+B_S(\beta;\mathcal{S})+B_D(\gamma;\mathcal{D}))$. $\mathcal{J}$ is a sparse matrix that computes joint location from personalized skull vertices with shape $B_S$ and trait $B_D$ components.
Different from FLAME~\cite{flame}, which regresses jaw joint from facial vertices, our joint regressor $\mathcal{J}$ is defined by experienced surgeons as the midpoint of mandible condyles~\cite{jawrig}, as condyle is the key anatomical structure involved in the jaw rotation and translation.
As mandible movement lies in a 6 Degree Of Freedom (DoF) manifold~\cite{jawrig}, containing both rotation and translation. We therefore model jaw movement using pose parameter $ \theta \in \mathbb{R}^6$. 
Thus the pose space is defined via a mandible joint $K=1$, plus an additional global head orientation. 

The personalized template $\mathbf{T}_p$ is a linear combination of general head template $\overline{\mathbf{T}}$, shape blend shape $B_S$, trait blend shape $B_D$, pose blend shape $B_P$ and expression blend shape $B_E$. 
It is defined as $\mathbf{T}_p({\beta}, {\gamma},{\theta}, {\phi})=\overline{\mathbf{T}}+B_S(\beta;\mathcal{S})+B_D(\gamma;\mathcal{D})+B_P(\theta;\mathcal{P})+B_E(\phi;\mathcal{E})$.
We define the general head template $\overline{\mathbf{T}}$ with an outer surface and inner skull geometry, including a mandible and a maxilla:
$\overline{\mathbf{T}}=\{\overline{\mathbf{T}}_{mdb},\overline{\mathbf{T}}_{mxl},\overline{\mathbf{T}}_f\}. $
% The mandible mesh $\overline{\mathbf{T}}_{mdb}$ has $9476$ vertices and $18948$ faces, while the maxilla mesh $\overline{\mathbf{T}}_{mxl}$ has $17575$ vertices and $35162$ faces. 
% We adopt the facial geometry from ~\cite{ict} and build the outer face surface mesh  $\mathbf{T}_f$ consisting of $10710$ vertices and $21212$ faces.
%
${B_S}$, $B_P$ and $B_E$ are defined similarly to FLAME. We refer readers to the supplemental page for more details.
% please refer to the supplemental page
%
We use 
$\{\mathcal{W},\mathcal{J},\beta,\gamma,\phi,\theta,\overline{\mathbf{T}},\mathcal{P},\mathcal{S},\mathcal{D},\mathcal{E}\}$ to parameterize $\mathcal{G}$ in the following paper.

In SCULPTOR, we aim to improve the model's ability to generate more characteristic faces with a physiologically correct constraint. However, simply using shared coefficients for representing facial and skeletal shape can hardly achieve this specific demand. Firstly, the parametric model focus on global face shape variation among the population, which is limited in capturing rare personality face features. 
Secondly, the \revise{skeletal modification} follows a certain anatomical distribution. 
% can not be unlimited. 
Therefore, to build a physiologically consistent facial variation space, we need to learn from real-world medical data to avoid artificial modifications that reduce the realism of face generation.

Our trait component improves the model's ability to generate more characteristic faces with a physiologically correct constraint. Specifically, ${B_D}$ is weighted linear combination of trait blend shape $\mathcal{D}$ and parameter $\gamma$, 
% computed from PCA through $\gamma$ , 
$B_D( \gamma;  {\mathcal D})={\mathcal D} \gamma$,
We define the trait blend shape as the variation of the corresponding vertices between skull and face.
% while manipulating the inner skull structure. 
Due to the high deformation complexity of the non-rigid soft tissue between skull and face, it is unlikely to model the inner muscles and fat efficiently while maintaining anatomically \revise{consistent}. Thus we adopt the plastic surgery CT data, which contains pre-surgery and post-surgery scans of human face. 
The CT data records the variation between skull-face states before and after the operation, enabling the skull-face correspondence variation and defining the skull-face variation space as a trait space. \revise{For more details, please refer to the supplemental page.}

% We firstly learn two parametric models from pre-surgery and post-surgery data respectively, denoted by average templates $\overline{\mathbf{T}}_{pre}^i$ and $\overline{\mathbf{T}}_{post}^i$. Then we extract the difference offset of skull-face as trait components as $d_i = \overline{\mathbf{T}}_{post}^i- \overline{\mathbf{T}}_{pre}^i$, which is thus able to conduct local characteristic edition on facial skeleton and 
% corresponding face shape. 

% Registration for inner outer combination
%%%%%%%%%%%%%%%%%%%%%%%%%%%%%%%%%%%%%%%%%%%%%%%%%%%
\subsection{Registration}
%%%%%%%%%%%%%%%%%%%%%%%%%%%%%%%%%%%%%%%%%%%%%%%%%%%
\label{sec:model_registration}
% qiu
% Bring them into the same topology
% Fixing topology with Laplacian smooth
% Fixing topology with manually corrected good mesh
% First the skull-face data from face
% Second, whether we need to use the NJU w/o skull?
% liyuwei
% what
% challenge, 
The first step in building a parametric model is to associate the template skull mesh and face mesh with all the individual skulls and faces in our LUCY dataset.
% extracted from clinical CT.% and multi-view 3D data. 

%%%%%%%%%%%%%%%%%%%%%%%%%%%%%%%%%%%%%%%%%%%%%%%%%%%
\noindent \textbf{Registration on skull.}
%%%%%%%%%%%%%%%%%%%%%%%%%%%%%%%%%%%%%%%%%%%%%%%%%%%%
The major difficulty is that the CT-generated skull mesh is incomplete around orbit and cheekbone, where the bones are too thin compared with CT image resolution, thus leaving numerous small holes in the mesh. 
To address this issue, we employ a semantic embedded deformation scheme based on \cite{xu2019flyfusion}, where we enforce larger mesh regularization for incomplete and noisy areas. 

% rigid align
First, the skull template $\overline{\mathbf{T}}_{s}=\{\overline{\mathbf{T}}_{mdb},\overline{\mathbf{T}}_{mxl}\}$ and CT skull $\mathbf{C}_s$ are roughly aligned using Procrustes rigid alignment on landmark correspondences. 
Then we use embedded deformation to recover skull details. 
We uniformly sample control nodes $x \in \mathcal{N}$ on the template surface with interval $\sigma$. Neighboring nodes are connected to form a graph, and then for each vertex $v$ in template $\overline{\mathbf{T}}_{s}$, the deformation is defined as:
\begin{equation}
    v' = \sum_{x \in \mathcal{N}} w(x,v) M v
\end{equation}
where $M$ is the transformation of node $x$, $w(\cdot)$ denotes the influence weight of node $x$ on $v$. 
We compute the weight using Radial Basis Function~\cite{rbf}, where a larger weight indicates a closer distance and stronger influence.
We optimize node deformation by minimizing the following energy term:
\begin{equation}
    E_{rskull} = E_{d} + \lambda_{l} E_{lmk} + \lambda_{r} E_{reg}
\end{equation}
where dense term $E_{d}$ enforces vertex alignment of the deformed template $\overline{\mathbf{T}}'_{s}$ and target CT skull $\mathbf{C}_s$ by 
\begin{equation}
    E_{d} = \lambda_d CD(\overline{\mathbf{T}}'_{s}, \mathbf{C}_s) + (1- \lambda_d)CD_n(\overline{\mathbf{T}}'_{s}, \mathbf{C}_s),
    \label{eqn:data}
\end{equation}
where $CD(\cdot)$ denotes the Chamfer Distance~\cite{chamfer} between two meshes.
% \begin{equation}
%     CD(\overline{\mathbf{T}}'_{s}, \mathbf{C}_s) = \frac{1}{|\overline{\mathbf{T}}'_{s}|}\sum_{v\in \overline{\mathbf{T}}'_{s}} \min_{u \in \mathbf{C}_s} \|v-u\|_2^2+\frac{1}{|\mathbf{C}_s|}\sum_{u \in \mathbf{C}_s}  \min_{v\in \overline{\mathbf{T}}'_{s}}\|u-v\|_2^2,
% \end{equation}
% where $u$,$v$ are the vertex that come from $\overline{\mathbf{T}}_{s}$, $\mathbf{C}_s$ respectively. 
%
$CD_n(\cdot)$ computes the angle between the corresponding vertex normal.
% \begin{equation}
%     CD_n(\overline{\mathbf{T}}'_{s}, \mathbf{C}_s) = \frac{1}{|\overline{\mathbf{T}}'_{s}|}
%     \sum_{v\in \overline{\mathbf{T}}'_{s}} 
%     \left(1-n_v\cdot n_{u_v}\right)
%     +\frac{1}{|\mathbf{C}_{s}|}\sum_{u \in \mathbf{C}_{s}}  \left(1-n_u\cdot n_{v_u}\right),
% \end{equation}
%
% where $n_v$ stands for the normal of vertex $v$ and $n_{u_v}$ represents the normal of vertex $u$, which is the corresponding vertex $v$ computed from Chamfer Distance.
$CD_n(\cdot)$ adds a normal penalty to prevent the template from fitting to the inner side of the maxilla, where the vertex normals are opposite.
$E_{lmk}$ is a sparse term to enforce landmark alignment; it computes the L2 distance between template landmark set ${L'}$ and target ${L}_S$:
\begin{equation}
    E_{lmk} = ||{L'} - {L_S}||^2_2.
    \label{eqn:lmk}
\end{equation}
Following \cite{li21piano} and \cite{xu2019flyfusion}, we adopt the as-rigid-as-possible motion regularization term $E_{reg}$ that enforces the neighbouring node to deform similarly. Please refer to \cite{xu2019flyfusion,newcombe2015dynamicfusion} for the complete formulation of this term.
Instead of using a uniform weight for all vertices, we set the weights for the orbit and nasal region 50 times larger than other parts. As a result, the vertices of these regions are more likely to deform in response to neighboring nodes, discarding erroneous correspondences caused by an incomplete target.

%%%%%%%%%%%%%%%%%%%%%%%%%%%%%%%%%%%%%%%%%%%%%%%%%%%
\noindent \textbf{Registration on face.}
%%%%%%%%%%%%%%%%%%%%%%%%%%%%%%%%%%%%%%%%%%%%%%%%%%%
%
Similarly, the face template $\overline{\mathbf{T}}_{f}$ and target face $\mathbf{C}_f$ are roughly aligned using Procrustes rigid alignment on landmark correspondences. Then we optimize template deformation by minimizing the mesh distance, landmark difference and a regularization term. It is defined as follows:
\begin{equation}
    E_{rface} = E_{d}(\overline{\mathbf{T}}_{f}, \mathbf{C}_f)+\lambda_{l} E_{lmk} + \lambda_{lap} E_{lap}.
\end{equation}
To maintain the well-defined topology of the template and suppress noise from original face data $\mathbf{C}_f$. 
We follow~\cite{flame} and adopt the discrete Laplacian regularization term, $E_{lap}$.
% \begin{equation}
%     E_{lap} = \frac{1}{N} \sum_{k=1}^N \frac{\lambda_v}{|\mathcal{N}(v)|}
%     \|\sum_{v_r \in \mathcal{N}(v)}(v_r-v)\|_2^2,
%     \label{eqn:lap}
% \end{equation}
% where $\mathcal{N}(v)$ represents the neighbors around vertex $v$ on the face template. $\lambda_v$ is the specific weight for each vertex.
%
Mesh distance term $E_{d}$ and landmark term $E_{lmk}$ are identical in Equ.\ref{eqn:data} and Equ.\ref{eqn:lmk}. 

%%%%% Training process on data%%%%%%%%%%%%%%%%%%%
\subsection{Parameter Learning}
\label{sec:parameter_learning}
%%%%%%%%%%%%%%%%%%%%%%%%%%%%%%%%%%%%%%%%%%%%%%%%%%%%
 
%%%%%%%%%%%%%%%%%%%%%%%%%%%%%%%%%%%%%%%%%%%%%%%%%%%%%%%%%%%%%%
After registration, the skulls and corresponding faces from LUCY have been aligned to the same topology as our general template.
We then set out to train the following model parameters $\{\overline{\mathbf{T}},\mathcal{S},\mathcal{D}, \mathcal{W}, \mathcal{P}\}$ similarly to previous works~\cite{flame, mano, smpl,li2022nimble}.
Though data in LUCY are captured under a neutral pose and expression, the definition of "neutral" varies from subject to subject, so we assume that each data has a minor pose and expression variation. 
Thus, in order to better disentangle shape from pose and expression, we adopt the expression basis from~\cite{ict} for disentanglement and utilize FaceScape~\cite{yang2020facescape} to learn pose-related parameters, i.e. skinning weight $\mathcal{W}$ and pose blend shape $\mathcal{P}$, so that we can better neutralize LUCY data. 

% \lyw{training overview}
We first learn the initial shape and trait parameters $\{\overline{\mathbf{T}}, \mathcal{S}, \mathcal{D}\}$ on LUCY, then train pose parameters $\{\mathcal{W}, \mathcal{P}\}$ on FaceScape. We iterate the learning on two data sources until convergence. 
%
% \lyw{parameter initialization}
%
We start with the general template $\tilde{\mathbf{T}}$ as the initial and initialize skinning weight with $\mathcal{W}$ which is transferred from FLAME~\cite{flame} via RBF kernel~\cite{rbf}. 
%
% We use Eqn.\ref{eqn:formulation} to compute the deformed model. 

\noindent \textbf{Learning on LUCY.}
We first use the LUCY dataset to train skeleton-consistent shape and trait components $\{\overline{\mathbf{T}}, \mathcal{S}, \mathcal{D}\}$. 
To this end, we need to solve for the neutral template $\mathbf{T}_p^i$ for each subject, $i$, as well as the corresponding minor pose parameter $\theta_i$ and expression parameter $\phi_i$. 
To disentangle the neutral facial shape from jaw pose and expressions, we train model parameters in two iterative steps: pose/expression and shape. In the first step, we fix shape parameters and optimize for pose and expression, and in the second step, we update parameters in the opposite way. This training process was carried out on the whole LUCY dataset without separating pre-surgery and post-surgery data.

Specifically, for each subject $i$, we compute the deformed model with optimized parameters and Eqn.\ref{eqn:formulation}, then minimize the following objective function:
\begin{equation}
    \begin{aligned}
        E_{L}(\theta_i, \phi_i, \mathbf{T}_p^i) =~ & \lambda_{vert} E_{vert} + \lambda_{edge} E_{edge} + \lambda_{lap} E_{lap} + \lambda_{sreg}E_{sreg},
    \end{aligned}
    \label{eqn:CT}
\end{equation}
where the data term $E_{vert}$ measures the euclidean distance between the deformed template mesh and the target registration, the edge term $E_{edge}$ computes the corresponding edge length difference. We refer readers to  \cite{smpl,flame} for further details. 
$E_{lap}$ is the discrete Laplacian~\cite{vertlap} term.
% in Eqn.\ref{eqn:lap}.
We apply it on $\mathbf{T}_p$ to force the vertices to preserve the original topology distribution so that it is robust to registration noise. 
$E_{sreg}$ is a shape regularizer that constrains the outer surface of $\mathbf{T}_p^i$ to be in the initial shape space by restraining the projected shape coefficients from being zero \cite{ict}. It is defined as follows:
\begin{equation}
    E_{sreg} = ||\mathbf{T}_{pf}^i \mathcal{\tilde{S}}^T||^2,
\end{equation}
where $\mathbf{T}_{pf}^i$ is the facial geometry of the neutral template of subject $i$, which is defined as the initial shape basis $\mathcal{\tilde{S}}$ from~\cite{ict}. 
We only add a shape regulator to the facial geometry here because~\cite{ict} only has a shape basis for the outer surface,

After iteratively learning for pose, expression and shape on LUCY, we obtain an estimated personal template $\mathbf{T}_p^i$ for both pre- and post- surgery data. 
Then we separate trait from shape component by first applying principal component analysis (PCA) on the post-surgery subset to obtain $\mathcal{S}$ and mean shape $\overline{\mathbf{T}}$.
Then, we compute $\mathcal{D}$ by performing PCA on the vertex offset of pre- and post-surgery data by $d_i = {{\mathbf{T}}}_{post}^i- {{\mathbf{T}}}_{pre}^i$ to model the trait component. 

\noindent \textbf{Learning on FaceScape.}
%%%%%%%%%%%%%%%%%%%%%%%%%%%%%%%%%%%%%%%%%%%%%%%%
After acquiring ${\overline{\mathbf{T}}}$, $\mathcal{{S}}$ and $\mathcal{{D}}$ from LUCY dataset,
we set out to learn pose-related parameters $\{\mathcal{W}, \mathcal{P}\}$. 
Firstly, we retopology FaceScape dataset to match our outer surface model. 
Then given registered facial geometry, we solve for the person-specific template $\mathbf{T}_p^i$ and compute joint position $J_p^i$ with $\mathcal{J}$. Then we optimize subject-specific parameters $\theta_i$ and $\gamma_i$ and global parameters  $\{\mathcal{W}, \mathcal{P}\}$. 
%
%%%%%%%%%%%%%%%%%%%%%%%%%%%%%%%%%%%%%%%%%%%%%%

The objective energy function is defined as follows:
\begin{equation}
    E_{F}(\beta_i, \theta_i, \gamma_i, \mathcal{W}, \mathcal{P}) = E_{vert} + \lambda_{col} E_{col} + E_{preg},
\end{equation} 
where $E_{vert}$ is the same in Eqn.~\ref{eqn:CT}, except that we only compute the L2 distance between the surface of the deformed model and the registered target, as FaceScape only contains face geometry. 
To avoid collision between skull and face, we use the collision term, $E_{col}$ in~\cite{Hasson_2019_CVPR} to penalize collision. 
To avoid overfitting to face, we add regularization on parameters:
\begin{equation}
    E_{preg} = \lambda_{\beta} E_{\beta}+\lambda_{\gamma} E_{\gamma} + \lambda_{\tilde{W}} E_{\tilde{W}} + \lambda_{p} E_{p}
\end{equation}
where $E_{\beta}$, $E_{\gamma}$ and $E_{\tilde{W}}$ are L-2 regularization terms on corresponding parameters. $E_{\tilde{W}}$ prevents optimized $\mathcal{W}$ becoming far away from initial $\tilde{\mathcal{W}}$.
Meanwhile, similar to~\cite{smpl}, we use Frobenius norm for $E_p$ to keep pose blend shape $\mathcal{P}$ sparse.

\noindent \textbf{Optimization Summary.}
We alternatively optimize the parameters on two different datasets. 
$\{\mathbf{\overline{T}}, {\mathcal{D}}$, ${\mathcal{S}}\}$ optimized from LUCY are then passed to FaceScape, the learning process on FaceScape then update skinning weight $\mathcal{W}$ and pose blend shape $\mathcal{P}$. 
The iteration goes on until convergence. 

\noindent\textbf{Appearance Modeling.}
Following \cite{qian2020html}, we model appearance with texture maps. Besides the texture maps from LUCY, we add extra online face texture assets from \cite{webscan3d}. We use a pre-defined UV-map to unify all the texture data and then perform principal component analysis directly on images. Thus $\mathcal{A}(\alpha)$ is able to produce a skin texture image given a random appearance parameter $\alpha$.

\begin{figure*}[t]
    \centering
\includegraphics[width=1\linewidth]{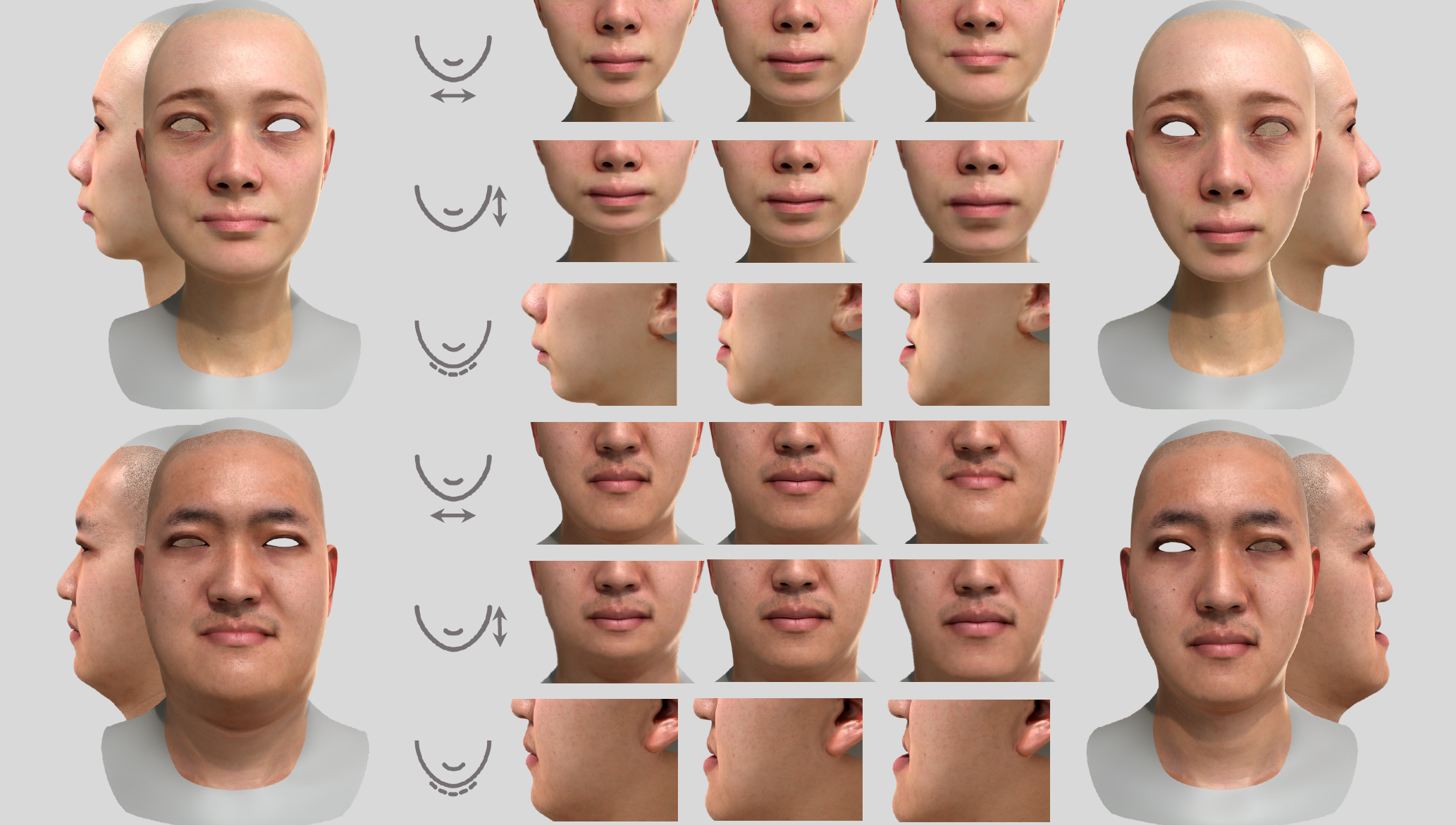}
    \caption{Performance of skeleton-driven characteristic face editing on one female (top row) and one male (bottom row) actors’ faces using the trait space in SCULPTOR. Each row of the partial enlargement displays the characteristic facial variations according to a representative trait component.}
    \label{fig:Gallery2}
\end{figure*}

\begin{figure*}[t]
    \centering
    \includegraphics[width=1\linewidth]{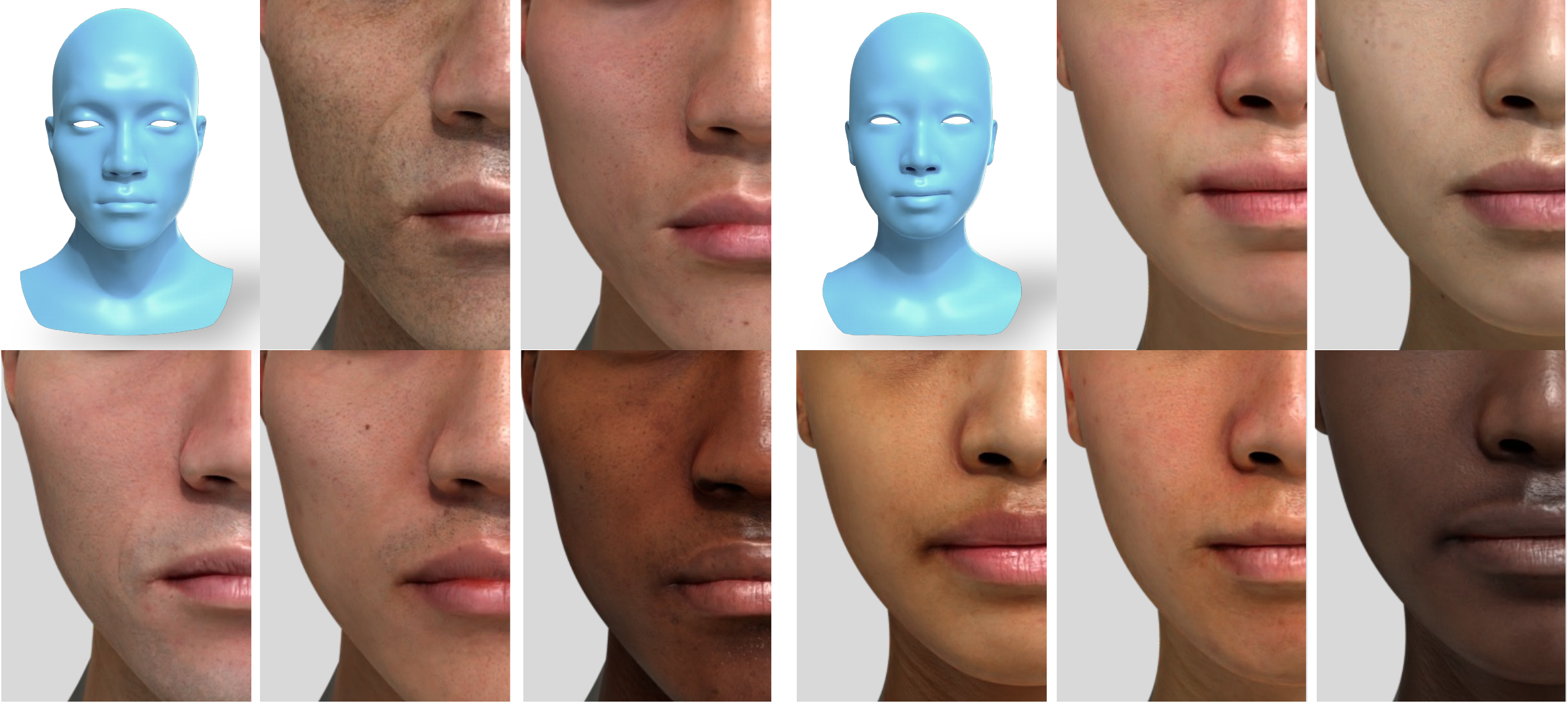}
    \caption{Examples of randomly generated facial appearance variations of SCULPTOR on a male and a female face respectively.  }
    \label{fig:Gallery3}
\end{figure*}

\subsection{Skeleton Consistent Generator}
\label{sec:rendering}
% %%%%%%%%%%%%%%%%%%%%%%%%%%%%%%%%%%%%%%%%%%%%%%%%%%%%%%%%%%%%%%%%%%%%%%%%%%%%

%\todo{============placeholder==========}
\noindent\textbf{Skeleton driven face generator.} 
For generating high-quality and realistic face appearance, most existing parametric face models achieve remarkable performance on face shape, expression detail and skin texture. But it is hard to locally carve the characteristic face skeletal details, such as the shape of jaw, the height of cheekbones, or curve of eyebrows. The SCULPTOR jointly models the correlation of skull and face as well as characteristic variance in a data-driven manner, thus deriving a skeleton-consistent face generator model. 
We show our procedural face generation process in Fig.~\ref{fig:pipeline}. We start with the SCULPTOR average template model, and randomly generate shape vector $\beta$ to make a variation on face shape. Then taking a randomly synthesized $\gamma$ vector as input, the model is further modified on skull local shape details, which add more features such as higher cheekbone, more curved jaw, or even mandibular protrusion to the generated face. For example, in Fig.~\ref{fig:pipeline}, we synthesize a man with a long square chin. Afterward, we continue adding facial appearance, pose, and expression parameters to generate a face with a widely opened mouth.
We render the final image through Blender~\cite{cycles} cycle render engine with high-resolution environment texture in the final generation process.

\noindent\textbf{Skeleton driven characteristic face editor.} 
As shown in Fig.~\ref{fig:pipeline}, limited by the ability of CT training data to present face local details, the face geometry directly generated from the proposed template is a bit over-smooth. We thus propose an alternative face generation pipeline for implementing skull edition from a real 3D face. We first use SCULPTOR model to fit with the 3D face in geometry, producing a smoothed model face. Then similarly to the previous section, by taking a randomly synthesized $\gamma$ vector as input, we add more trait structure details to the skull model.
Meanwhile, the face geometry is also changed accordingly. Finally, we add the personal skin offset back to the curved face geometry and generate a new face with characterized face contour and appearance details. The performance of this generator is shown in Fig. ~\ref{fig:Gallery2}.

%% file: sections/application.tex
\section{Application}
\label{sec:application}

%%%%%%%%%%%%%%%%%%%%%%%%%%%%%%%%%%%%%%%%%%%%%%%%%%%%%%%%%%%%%%%%
\subsection{Archaeological Skeletal Facial Completion}
%%%%%%%%%%%%%%%%%%%%%%%%%%%%%%%%%%%%%%%%%%%%%%%%%%%%%%%%%%%%%%%%
% \todo{======placeholder==========}

During archaeological excavation, archaeologists rarely recover complete evidence. Especially human skeletons, during the long history and complex influence of natural factors, organic materials such as human bones are more challenging to preserve intact than inorganic relics. As one of the most mysterious ancient females, part of Ava's remains was found during the excavation in Caithness, Scotland, 1987. This 4200-year-old young lady only left her mandible, and her DNA indicates that she likely had bronze skin, brown eyes and black hair~\cite{ava}. Using Ava's mandible and teeth as a basis, archaeologists have commissioned forensic artists to conduct the original facial reconstruction of Ava. Using SCULPTOR, we have more statistical evidence and flexibility to complete her facial skeleton and infer different facial appearances.

% 1.	How to use parametric model fit the mandible of Ava. Using the parametric model to regress basis weight from maxilla, and propagate these weight to form a corresponding lower jaw. 

% 我们用300个shape component的模型通过E_ava去优化参数beta来得到ava的上颌骨预测
We first intend to use the maxilla part of the proposed SCULPTOR model to fit with Ava's maxilla. Specially, we adopt our model with 300 shape components to optimize shape parameter $\beta$ with energy function: $E_{Ava} = E_{vert} + \lambda_\beta E_{\beta} $, where $E_{vert}$ is the mean vertex error in Eqn.~\ref{eqn:CT} while $E_{\beta}$ regularizes the shape parameter. 

% \begin{equation}
%     E_{mxl}=\|\mathcal{G}(\theta, \beta,0,0)_{mxl}-T_{mxl}^{Ava}\|_2^2
% \end{equation} 

%  在$E_{mxl}$中, $T_{mxl}^{Ava}$是通过对原ava上颌骨registration  后得到的aligned Ava上颌骨的模型.为了防止模型对aligned上颌骨overfit, 我们增加了一个对beta的L2 regulation. 我们优化beta来得到与aligned上颌骨匹配的上颌骨.
% 3.	How to generate face mesh according to the skull we synthesized.
% 因为我们的模型是上颌, 下颌, 脸一起训练的, 所以他们共享一套beta参数, 于是我们可以得到上颌骨相对应的下颌骨与脸.
%  In the above equation,  $T_{mxl}^{Ava}$ denotes the mesh geometry of Ava's maxilla, which is computed through the pipeline in Sec \ref{sec:model_registration}. 
 We optimize $\beta$ by regressing the maxilla part of our parametric model. %to approach maxilla%$T_{mxl}^{Ava}$. 
 Then, the optimized shape coefficient $\beta$ is applied to model components related to the maxilla, mandible and face geometry to generate the corresponding face and mandible.% according to Ava's maxilla.
 %With concatenated shape components, we generate the corresponding face and mandible through the mandible, maxilla and face shape components share the same weight.
%
% 2.	How to synthesis different lower jaw based on the result of point 1. (using diversity basis)
% 为了生成不同下颌骨的ava脸, 我们使用300个 shape components与50个diverstiy components 来优化beta 与gamma通过E_{Ava}. 我们首先 randomly init diversity weight, gamma0初始化下颌的形状, 然后整体优化beta 与 gamma使得生成的上颌骨趋近于ava的aligned 上颌骨. 为了避免生成的下颌骨远离初始生成的下颌骨, 我们增加了$E_{\gamma_{0}}=\|\gamma_{0}-\gamma\|_2^2$让优化的gamma不远离gamma0. 于是生成各式各样的ava的脸with不同的下颌骨.

% 4.	How to add texture to face.

To generate various of Ava with the fixed maxilla and different mandibles, we further apply our model with 300 shape components and 50 trait components to jointly optimize $\beta$ and $\gamma$ on with energy function: $E_{Ava} = E_{vert} + \lambda_\beta E_{\beta} + \lambda_{\gamma_0} E_{\gamma_{0}}$.
% \end{equation}
where $\gamma_{0}$ is a randomly initialized vector that brings a creative and anatomically consistent modification to Ava's skeletal structure. The effect of the trait components mainly focuses on mandible shape, but there still exists a bit of influence on the maxilla. We thus need to slightly refine the model by minimizing the energy function $E_{Ava}$. We additionally define $E_{\gamma_{0}}=\|\gamma_{0}-\gamma\|_2^2$ to keep $\gamma$ as close to the initialization of $\gamma_{0}$ as possible when we optimize $\beta$ and $\gamma$.

%%%%%%%%%%%%%%%%%%%%%%%%%%%%%%%%%%%%%%%%%%%%%%%%%%%%%%%%%%%%%%%%
\subsection{Character Fusion}
%%%%%%%%%%%%%%%%%%%%%%%%%%%%%%%%%%%%%%%%%%%%%%%%%%%%%%%%%%%%%%%%

% \todo{============placeholder==========}

For generating more diverse and realistic faces, we intend to exchange the upper and lower facial skeleton structures of two different actors instead of simply adjusting the coefficients of the parametric trait components in SCULPTOR. 

% 1.	Regress the two skulls using SKAVENGER ( don’t talk about the pose correction)

% 2.	How to exchange two lower jaws. (In case they are with different size and position of the jaw connection point)

% 我们用a的上颌与b的下颌组成了一组新的skull, {T}_{new}. 然后我们用模型\mathcal{G}(\theta, \beta,\gamma,0)_{skull}|_2^2优化beta, gamma, theta.因为我们的模型是上颌, 下颌, 脸一起训练的, 所以他们共享一套beta参数, 所以我们获得了新骨头下的脸. 为了transfer另一个人的表情与动作, 我们用模型去fit他的参数后把他动作用模型transfer到我们生成的人上.

Specially, we first generate a new skull using the maxilla of actor $a$ and mandible of actor $b$  as  $\mathbf{T}_{new}=[\mathbf{T}_{mxl}^a,\mathbf{T}_{mdb}^b]$. Then we optimize the pose parameter $\theta$, shape parameter $\beta$ and trait parameter $\gamma$ to minimize:

\begin{equation}
    E_{fuse} = E_{vert}+ \lambda_\beta E_{\beta} + \lambda_\gamma E_{\gamma} + \lambda_{sim} E_{sim}
\end{equation}
% 
% \begin{equation}
%     E_{fuse} = \|\mathbf{T}_{new}-\mathcal{G}(\theta, \beta,\gamma,\mathbf{0})_{skull}\|_2^2+ \lambda_\beta E_{\beta} + \lambda_\gamma E_{\gamma} + \lambda_{sim} E_{sim}
% \end{equation}
where $E_{\gamma}$ regularizes the trait parameter and $E_{sim}$ is designed to keep the generated face as close as possible to the original faces.
Since face shape, trait components, skeleton shape and trait components share the weight parameters. 
% We can then infer the face with a neutral pose from the combined skeleton by setting pose parameter $\theta=\mathbf{0}$.%, thus %update $\mathcal{G}(\theta, \beta,\gamma,\mathbf{0})_{skull}$ as $\mathcal{G}(\mathbf{0}, \beta,\gamma,\mathbf{0})_{skull}$. the combined face is computed using $\mathcal{G}(\mathbf{0}, \beta,\gamma,\mathbf{0})_{face}$. 
After fixing the skull model, for generating more pose and expression variation to the facial appearance, we can use pose $\theta_s$ and expression parameters $\phi_s$ to control the facial pose and expression based on the newly generated face.% the combined face model as $\mathcal{G}(\theta_s, \beta,\gamma,\phi_s)_{face}$.

% further transfer the expression and pose from a third face by estimating the pose and expression parameter using the whole SCULPTOR model parameters. We optimize pose , shape parameter $\beta_s$, diversity parameter $\gamma_s$ and expression parameter  by fitting the face geometry  to the source facial mesh. And finally apply the parameters to the combined face model as $\mathcal{G}(\theta_s, \beta,\gamma,\phi_s)_{face}$.

% 3.	Based on the new skeleton, how to build corresponding facial surface mesh.

% \todo{4.	How to add texture.}
%  placeholder placeholder placeholder placeholder placeholder placeholder placeholder placeholder placeholder placeholder placeholder placeholder placeholder placeholder placeholder placeholder placeholder placeholder placeholder placeholder placeholder placeholder placeholder placeholder placeholder placeholder placeholder placeholder placeholder placeholder placeholder placeholder placeholder placeholder placeholder placeholder placeholder placeholder placeholder placeholder placeholder placeholder placeholder placeholder placeholder placeholder placeholder placeholder placeholder placeholder placeholder placeholder placeholder placeholder 

% \todo{============placeholder==========}

%%%%%%%%%%%%%%%%%%%%%%%%%%%%%%%%%%%%%%%%%%%%%%%%%%%%%%%%%%%%%%%%
\subsection{Skull Inference from Image}
%%%%%%%%%%%%%%%%%%%%%%%%%%%%%%%%%%%%%%%%%%%%%%%%%%%%%%%%%%%%%%%%

Similar to other parametric models~\cite{flame,mano,smpl,li2022nimble}, our model is easy to apply to skull inference tasks from images. 
We treat our model as a differentiable layer that takes pose $\theta$, shape $\beta$, trait $\gamma$, expression $\phi$ and appearance $\alpha$ parameters as input and directly outputs a 3D head with inner skull and outer skin geometry as well as high-quality facial textures. 
For skull inference from images, we build upon the DECA network structure \cite{feng2021learning_deca}, which takes 2D in-the-wild images as input and outputs a 3D animation face. 
We replace FLAME model in the DECA training loop with our model and add another regression branch for the trait components, then we train the network to regress model parameters and camera extrinsic on the FFHQ dataset \cite{karras2019style}.
Similarly, we use landmark re-projection loss, photometric loss, identity loss, shape consistency loss, parameter regularization loss and detail reconstruction loss to train the network in an analysis-by-synthesis manner. We refer readers to \cite{feng2021learning_deca} for details. 
Note that our anatomical landmark definition deviates from the general facial landmark, so we manually redefine corresponding landmarks in our template facial geometry to match the dataset annotation.

\begin{figure*}[t]
    \centering
    \includegraphics[width=1\linewidth]{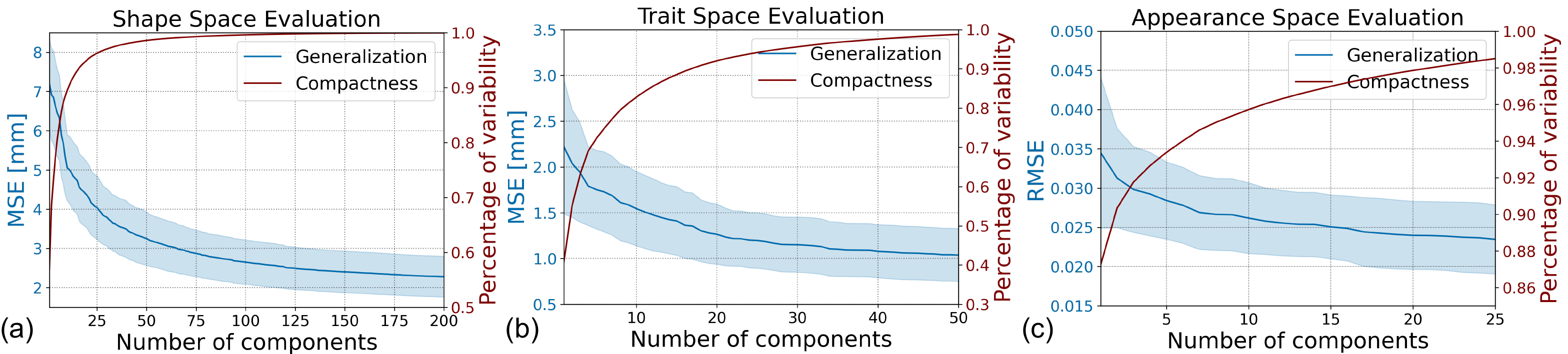}
    \caption{Model evaluation on compactness and generalization. From
     left to the right: (a) Shape space (b) Trait space (c) Appearance space.}
    \label{fig:PCeval}
\end{figure*}

%%%%%%%%%%%%%%%%%%%%%%%%%%%%%%%%%%%%%%%%%%%%%%%%%%%%%%%%%%%%%%%%
% \subsection{Same character over time. }
\subsection{Face Generation with Lipo Level Change Effect}
\label{sec:bmi}
%%%%%%%%%%%%%%%%%%%%%%%%%%%%%%%%%%%%%%%%%%%%%%%%%%%%%%%%%%%%%%%%
% puffiness change
To create the lipo level change effect, we start with multiple images of the same subject with varying levels of lipo. Then we infer the skull and head shape with neutral pose and expression from each image.
Since we assume that the 3D offsets between the face and skull do not vary much between subjects throughout our model learning, i.e., a constant lipo level distribution, the predicted skull shape is different for each image. However, in reality, the skull shape of an adult is invariant to body weight growth or loss. 
As a result, we use the prediction with the minimum fitting error as the subject's anticipated skull since it most closely matches the lipo-level distribution of our dataset. We denote it as $\hat{T}_{skull}$ and the corresponding neutral head as $\hat{T}_{f}$.
To create a realistic lipo level transition effect, \cite{PhaceICHIM} proposed to use a hand-painted "lipo map" to specify which areas of the head are more prone to body weight accumulation. Similarly, we automatically create a person-specific "lipo map" with $\hat{T}_{skull}$ and all predicted head shapes $\mathcal{T}_{f}$. It is defined by vertex weight on the face mesh:
\begin{equation}
    \omega_i = \frac{1}{N} \sum_{v\in {T}^j_{f}} \min_{u\in \hat{T}_{skull}}|| v  - u||^2_2,
\end{equation}
where $i$ denotes the $i^{th}$ vertex on the face surface mesh, $N=|\mathcal{T}_{f}|$ is the face mesh count and $T^j_{f} \in \mathcal{T}_{f}$ denotes the $j^{th}$ face mesh. Larger weight specifies that the vertex is more likely to shift when lipo level changes, namely more variant to weight growth and loss. We also regulate $\omega$ by maximum normalization for the following computation. See Fig.~\ref{fig:BMI} for the visualization of personal lipo maps. 
Next, we apply PCA on the offset of all head shapes with the neutral one $\mathcal{T}_{f} - \hat{T}_{f}$ and use the principal components $\mathcal{B}$ and coefficients $\delta$ as an initial person-specific puffiness component. Although directly modifying $\delta$ achieves a smooth lipo level change effect, the results may suffer from unnatural deformation and visual artifacts around eyes and ears as these regions are expected to be invariant to weight change. Thus, we further optimize $\mathcal{B}$ and $\delta$ such that the deformation matches the personal lipo map. 
Denoting the recovered head mesh using lipo components as $T_{lipo} = \hat{T}_{f} + \mathcal{B} \delta$, the optimization energy is defined as:
\begin{equation}
    E_{lipo} =  \omega(\mathcal{G}_{lipo} - \mathcal{T}_{f}) + (1-\omega)(\mathcal{G}_{lipo} - \hat{T}_{f}).
\end{equation}
$E_{lipo}$ constraints deformation to align with inferred face $\mathcal{T}_{f}$ when lipo map has larger values and reduces the deformation effect for regions with a smaller weight. 
Finally, we perform another PCA on $\mathcal{G}_{puff} - \hat{T}_{f}$, so we can interpolate and extrapolate anatomically \revise{consistent} faces with various puffiness levels by modifying $\delta$. The results are shown in Fig.~\ref{fig:BMI}.

\begin{table}[t]
\centering
\caption{Quantitative results for skull fitting performance. We evaluate Mean Squared Errors in millimeter on pre- and post-surgery test scans.
SCULPTOR-SIMPLE stands for the simplified version which only models shape, while SCULPTOR models both shape and trait components.   
% SCULPTOR-SIMPLE is trained to have 144 shape components by treating 72 scan pairs as different scans while SCULPTOR extracts both shape and trait components from 72 pairs of pre- and post-surgery scans.
}
\begin{tabular}{c|c|c}
\hline
Model & pre-surgery   & post-surgery  \\
\hline
SCULPTOR-SIMPLE   &2.01  &  2.04   \\
SCULPTOR &1.77  &  1.77 \\
\hline
\end{tabular}
\label{tab:skull_inference}
\end{table}

%%%%%%%%%%%%%%%%%%%%%%%%%%%%%%%%%%%%%%%%%%%%%%%%%%%%%%%%%%%%%%%%
% \subsection{Physical simulation - movements and virtual plastic surgery }
\subsection{Facial Animations}
%%%%%%%%%%%%%%%%%%%%%%%%%%%%%%%%%%%%%%%%%%%%%%%%%%%%%%%%%%%%%%%%
To further demonstrate the advantage of modeling the inner structure, we apply physical simulation to our model to include secondary deformation under dynamic pose and external force. We show results of the skin deformation during a rapid head-shaking and under a fist punch.
Similar to \cite{kozlov2017enriching}, given a generated 3D head with inner skull and outer skin, we create a simple volume of tissue between them. We use tetrahedral mesh to represent the soft tissue and have artists manually tune the material parameters for soft-body dynamics. Using collision detection, we additionally ensure that no skin penetrates the skull.
To compare with models without inner structure, we consider the whole head as a whole volume by omitting the inner skull and rerunning the simulations again. 
Visual results are shown in Fig.~\ref{fig:PunchFace}. Please also see the accompanying video for the dynamic sequence.

% \subsection{Physical simulation  }
% \paragraph{movements and virtual plastic surgery } 

% \begin{figure*}[t]
% \centering
% \includegraphics[width=0.9\linewidth]{figures/gallery.pdf}
% \caption{gallery}
% \label{fig:gallery}
% \end{figure*}

%\paragraph{Diversity Transfer}
% From a to b
% surgery transfer

% \begin{figure}
%     \centering
%     \includegraphics[width=0.2\linewidth]{figures/surgery_patient.pdf}
%     \caption{Diversity Transfer}
%     \label{fig:Diversity_Transfer}
% \end{figure}

% \section{Stylized Facial Asset Generation}

% \subsection{StyleSDF basics and how to add skeleton as constraints. }
% \paragraph{BMI stimulation}

% \subsection{Initial results.  }
% \paragraph{BMI stimulation}

%% file: sections/experiment.tex
%%%%%%%%%%%%%%%%%%%%%%%%%%%%%%%%%%%%%%%%%%%%%%%%%%%%%%%%%%%%%
\section{Experiments}
\label{sec:experiment}
%%%%%%%%%%%%%%%%%%%%%%%%%%%%%%%%%%%%%%%%%%%%%%%%%%%%%%%%%%%%%
    
\begin{figure}[t]
    \centering
    \includegraphics[width=\linewidth]{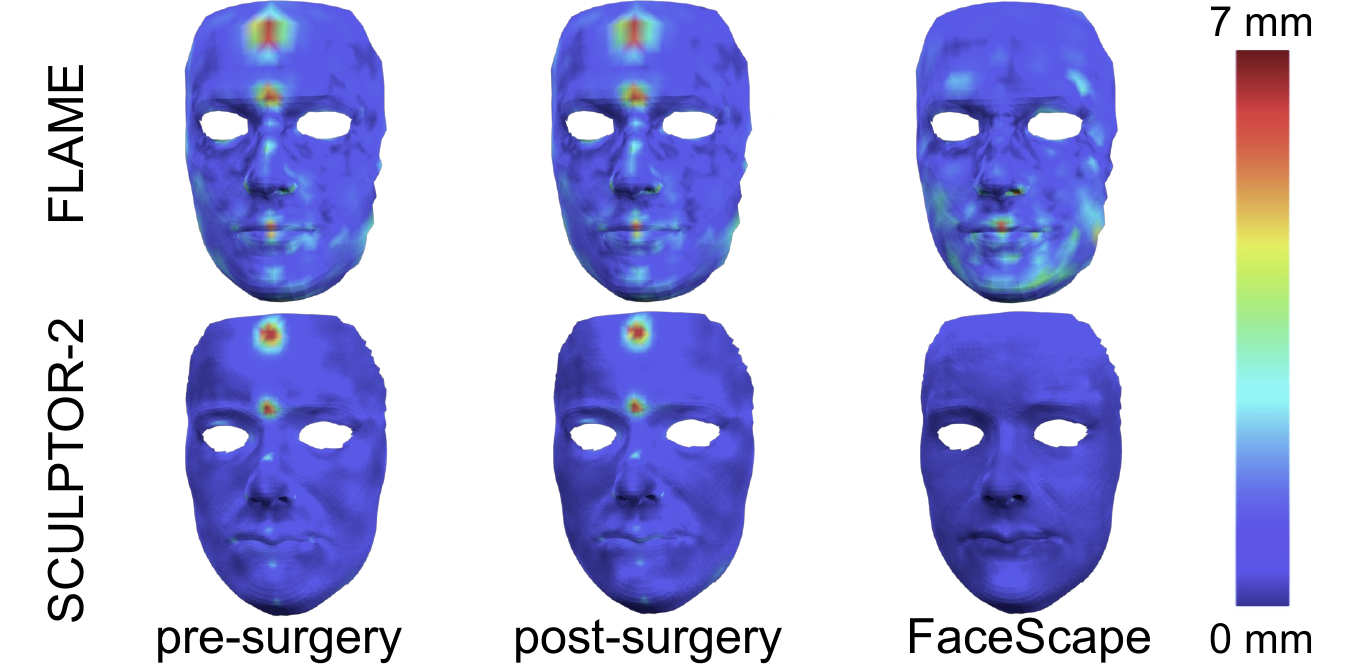}
    \caption{\revise{Mean squared per-vertex error between FLAME and SCULPTOR-2, measured on 
    pre-surgery, post-surgery and FaceScape test set.}}
    \label{fig:qualitative_recon}
\end{figure}

\begin{figure*}[t]
\centering
\includegraphics[width=1\linewidth]{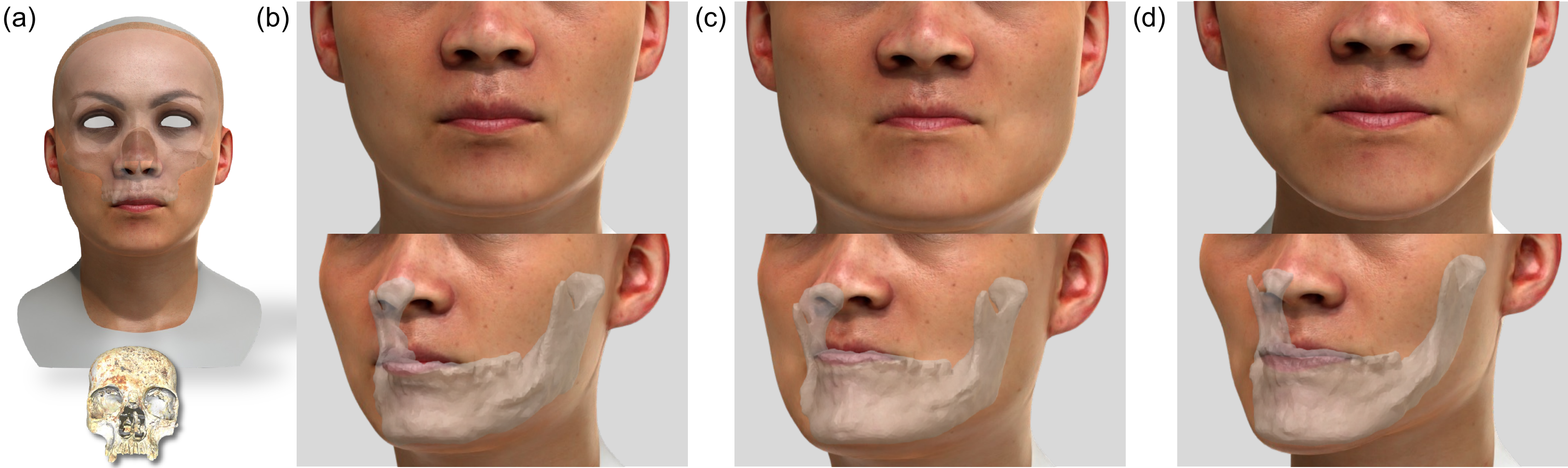}
\caption{Archaeological Skeletal Facial Completion. (a) The original maxilla of Ava and face was generated using SCULPTOR without trait components. 
Rendered skull image from Open Virtual World (Sketchfab). (b)-(d) Characteristic face generations with respect to Ava's maxilla by varying trait parameters in SCULPTOR. }
\label{fig:ava}
\end{figure*}

%%%%%%%%%%%%%%%%%%%%%%%%%%%%%%%%%%%%%%%%%%%%%%%%%%%%%%%%%%%%%
\subsection{Implementation Details}
\label{sec:implementation_details}
%%%%%%%%%%%%%%%%%%%%%%%%%%%%%%%%%%%%%%%%%%%%%%%%%%%%%%%%%%%%%

% For registration, we set $\lambda_l$ to 0.01 and $\lambda_r$ to 0.001 in practice. The embedded deformation is performed 4 times with node interval $\sigma$ ranging from 50 mm to 2 mm. The whole registration takes approximately 2 minutes per data on an NVIDIA GTX TITAN X GPU. 
In registration on the maxilla, we set $\lambda_d$ to 0.5  as small holes and fragments exist in the raw maxilla mesh, removing the effect of foldovers. $\lambda_l$ to 0.005 and $\lambda_r$ to 0.0005 in practice, enforcing our template is able to be aligned to the biological landmarks.  The maxilla registration takes approximately 5 minutes per data. 
In registration on the mandible, we set $\lambda_d$ to 1.0, $\lambda_l$ to 0.01 and $\lambda_r$ to 0.001 since there are no fragments in raw mandible mesh. 
For face registration, as we focus on the topology of the face that can still be maintained during registration, we set $\lambda_l$ to 0.03, $\lambda_{lap}$ to 10 and $\lambda_{d}$ to 0.5.
Each embedded deformation is performed 4 times with node interval $\sigma$ ranging from 50 mm to 2 mm. The mandible registration takes approximately 5, 2, 3 minutes per data for maxilla, mandible and face, respectively.

During parameter training, in the stage of learning from CT data, keeping $\lambda_{vert}=1.0$ and $\lambda_{sreg}=1.0$ in the whole process. We first set the $\lambda_{edge}=1.0$ to update the initial pose without any shape in the first iteration and keep the $\lambda_{edge}=0.0$ in the following optimization. We set $\lambda_{lap}=15.0$ to maintain the topology of the face and skull. In the FaceScape learning stage, we set $\lambda_{col}=50.0$ to strictly avoid the skull penetrating the face. For the rest parameters, we set $\lambda_{\beta}=0.1$, $\lambda_{\gamma}=0.1$, $\lambda_{\tilde{W}}=0.2$ and $\lambda_{p}=2.0$. We iteratively optimize the parameters until the model has converged through the optimization process with Limited-memory BFGS optimizer in PyTorch. We optimize all the parameters on an NVIDIA GTX TITAN X GPU.

\begin{table}[t]
    \centering
    \caption{Comparison with other face parametric models. We report mean squared error in millimeters \revise{in the first three columns, and chamfer distance (in millimeters) in 3DRFE dataset (the last column)} on facial mesh fitting tasks. We use 300 shape components for FLAME, 300 shape components for SCULPTOR-1 and 228 shape components, 72 trait components for SCULPTOR-2.}
    \begin{tabular}{c|c|c|c|c}
    \hline
    Model & pre-  & post- & FaceScape & \revise{3DRFE}\\
    \hline
    FLAME       & 1.58   & 1.60  &1.63  & \revise{2.79} \\
    SCULPTOR-1  &1.51   & 1.54  &0.66 &  \revise{2.86}\\
    SCULPTOR-2  &1.36  & 1.41  &0.67 & \revise{2.78}\\
    \hline
    \end{tabular}
    \label{tab:QuantiFaceRecon}
\end{table}

%%%%%%%%%%%%%%%%%%%%%%%%%%%%%%%%%%%%%%%%%%%%%%%%%%%%%%%%%%%%%
\subsection{Model Evaluation}
\label{sec:exp_eval}
%%%%%%%%%%%%%%%%%%%%%%%%%%%%%%%%%%%%%%%%%%%%%%%%%%%%%%%%%%%%%

% Compactness and generalization are two main metrics to measure the quality of statistical model~\cite{statistical_model_compact_general}. 
We use two metrics, compactness and generalization, to evaluate the quality of our statistical model. The shape space is learned from 72 post-surgery patient scans and 400 face scans with fitted skulls from~\cite{yang2020facescape}, and the trait space is computed from 72 pairs pre-surgery and post-surgery CT scans. Finally, the appearance space is trained with 126 high-resolution texture maps, 48 of them are online assets from \cite{webscan3d}. 

%%%%%%%%%%%%%%%%%%%%%%%%%%%%%%%%%%%%%%%%%%%%%%%%%%%%%%%%%%%%%
\paragraph{Compactness.}
%%%%%%%%%%%%%%%%%%%%%%%%%%%%%%%%%%%%%%%%%%%%%%%%%%%%%%%%%%%%%
% Fig 画了 SCULPTOR的shape与diversity space的 compactness. 这些曲线表达了在用不同数量的主成分展现了训练数据中的variance. 对于shape space, 我们画了前200个主成分. 我们的shape space是通过72对术前术后CT scan以及400来自于南京大学数据的neutral face与fitted骨骼. 在figure~\ref{fig:PCeval}(a)里, 我们发现在用13个主成分时就已经覆盖了90%的shape space. 与此同时, 25个主成分能覆盖超过95%的完整的shape space.在figure~\ref{fig:PCeval}(b)里, 我们画了前50个主成分, diversity space是通过72对术前术后CT scan计算得来. Diversity 的curve 展示了在第10, 20, 40的时候, 能够分别表达84%, 92%, 97%的整个diversity space. 

% Since in the CT scans one skull is only related with a single face geometry, we perform our shape learning with additional face data augmentation.
Fig.~\ref{fig:PCeval} (a), (b) (c) (dark red curve) plot the compactness of the SCULPTOR shape, pose and appearance space, respectively. These curves depict the variance in the training data captured by a varying number of principal components. In Fig.~\ref{fig:PCeval}(a), the curve implies that with the first 25 principle components, the shape space is able to cover 95\% of the entire space. 
Meanwhile, 200 principle components are able to express nearly the entire space. 
In trait space, the first 50 principle components are plotted with a dark red curve in Fig.~\ref{fig:PCeval}(b). It shows that the first 10 principle components achieve over 80\% of the trait space, while 50 components are able to cover nearly 98\% of the trait space.
The appearance space is also built with high compactness, where the first 10 principle components achieve 96\% of the full space while 25 components are able to cover over 98\% of the appearance space.

% STD table
% \begin{table}[t]
%     \centering
%     \begin{tabular}{c|c|c|c}
%     \hline
%     Model & pre-surgery  & post-surgery & FaceScape \\
%     \hline
%     FLAME       & 1.58 $\pm$ 0.13  & 1.60 $\pm$ 0.10  &1.63 $\pm$ 0.09 \\
%     SCULPTOR-1  &1.51 $\pm$ 0.18  & 1.54 $\pm$ 0.29  &0.66 $\pm$ 0.71 \\
%     SCULPTOR-2  &1.36 $\pm$ 0.18  & 1.41 $\pm$ 0.28  &0.67 $\pm$ 0.71 \\
%     \hline
%     \end{tabular}
%     \caption{Comparison with other face parametric models. We report mean squared error and corresponding standard deviation in millimeters on skull and facial mesh fitting tasks. We use 300 shape components for FLAME, 300 shape components for SCULPTOR-1 and 228 shape components, 72 trait components for SCULPTOR-2.}
%     \label{tab:QuantiFaceRecon}
% \end{table}

\begin{table}[t]
    \centering
    \caption{\revise{Quantitative reconstruction results in missing mandible or maxilla tasks. We report hausdorff distance in millimeters. }}
    \begin{tabular}{c|cc|cc}
\hline
           & \multicolumn{2}{l|}{Missing maxilla} & \multicolumn{2}{l}{Missing mandible} \\ \hline
           & \multicolumn{1}{l|}{pre-}   & post-  & \multicolumn{1}{l|}{pre-}   & post-  \\ \hline
SCULPTOR-1 & \multicolumn{1}{l|}{2.304}  & 2.235  & \multicolumn{1}{l|}{2.668}  & 2.421  \\
SCULPTOR-2 & \multicolumn{1}{l|}{2.149}  & 2.156  & \multicolumn{1}{l|}{2.259}  & 2.193  \\ \hline
    \end{tabular}
    \label{tab:MissingJaw}
\end{table}

%%%%%%%%%%%%%%%%%%%%%%%%%%%%%%%%%%%%%%%%%%%%%%%%%%%%%%%%%%%%%
\paragraph{Generalization.}
%%%%%%%%%%%%%%%%%%%%%%%%%%%%%%%%%%%%%%%%%%%%%%%%%%%%%%%%%%%%%
% 为了研究在有限的shape space 下的generalization ability, 我们用57个没见过的CT scan data 的评估方式在472例CT post scan 与fitted facescape组成的data上做评估.我们在展示图中的mean squared error and the standard deviation用毫米为单位. 在Figure~\ref{fig:PCeval}(a)中, 蓝色曲线代表了generalization curve的generalization误差. 随着主成分数目的增加, mean squared error随之下降, 在使用50, 100个主成分时 mean squared error分别下降到3.3mm 与 2.7 mm, 是在monotony的下降.
Fig.~\ref{fig:PCeval} (a), (b) (c) (blue curve) plot the evaluation result on the generalization ability of the SCULPTOR shape, pose and appearance space, respectively.
We use an additional 45 CT head image data  from the archived medical records outside the LUCY training dataset to evaluate shape-space generalization. 
In Fig.~\ref{fig:PCeval}(a), the blue curve depicts shape space generalization error.
The shape regression error is computed via the mean squared vertex error(MSVE), and the standard deviation is denoted in millimeters (mm).
The vertex error decreases monotonically on the test shape with respect to the increasing number of principle components. The vertex error curve achieves lower than 3.25 mm and 2.33 mm using 50 and 175 principle components, respectively.
In the trait space generalization study, aiming to generalize unseen traits under the limited principle components, the leave-one-out strategy is adopted for the 72 pairs of pre-surgery and post-surgery CT in the training data. In Fig.~\ref{fig:PCeval}(b) the blue curve evaluates the reconstruction error varying with an increasing number of involved principle components. With the increasing principle components, the mean squared error decreases to approximately 1.1 mm in using 50 principle components. Similarly, the appearance space also gives out the decreasing error curve with the number of components increasing.

% 为了研究在有限的diversity space 下的generalization ability, 换句话说时对没见过的diversity的的估计.  我们用leave one out 的评估方式对72对术前术后的CT scan数据做评估.我们在展示图中的mean squared error and the standard deviation用毫米为单位. 在Figure~\ref{fig:PCeval}(a)中, 蓝色曲线代表了generalization curve的generalization误差. 随着主成分数目的增加, mean squared error随之下降, 在使用50, 100个主成分时 mean squared error分别下降到0.8mm 与 0.55 mm. with the increasing principle components, the mean squared error decreases to approximately at 1.0 mm in using 50 principle components. Compared to shape space, the high standard deviation implies the complexity of diversity space
\begin{figure}[t]
\centering
\includegraphics[width=0.95\linewidth]{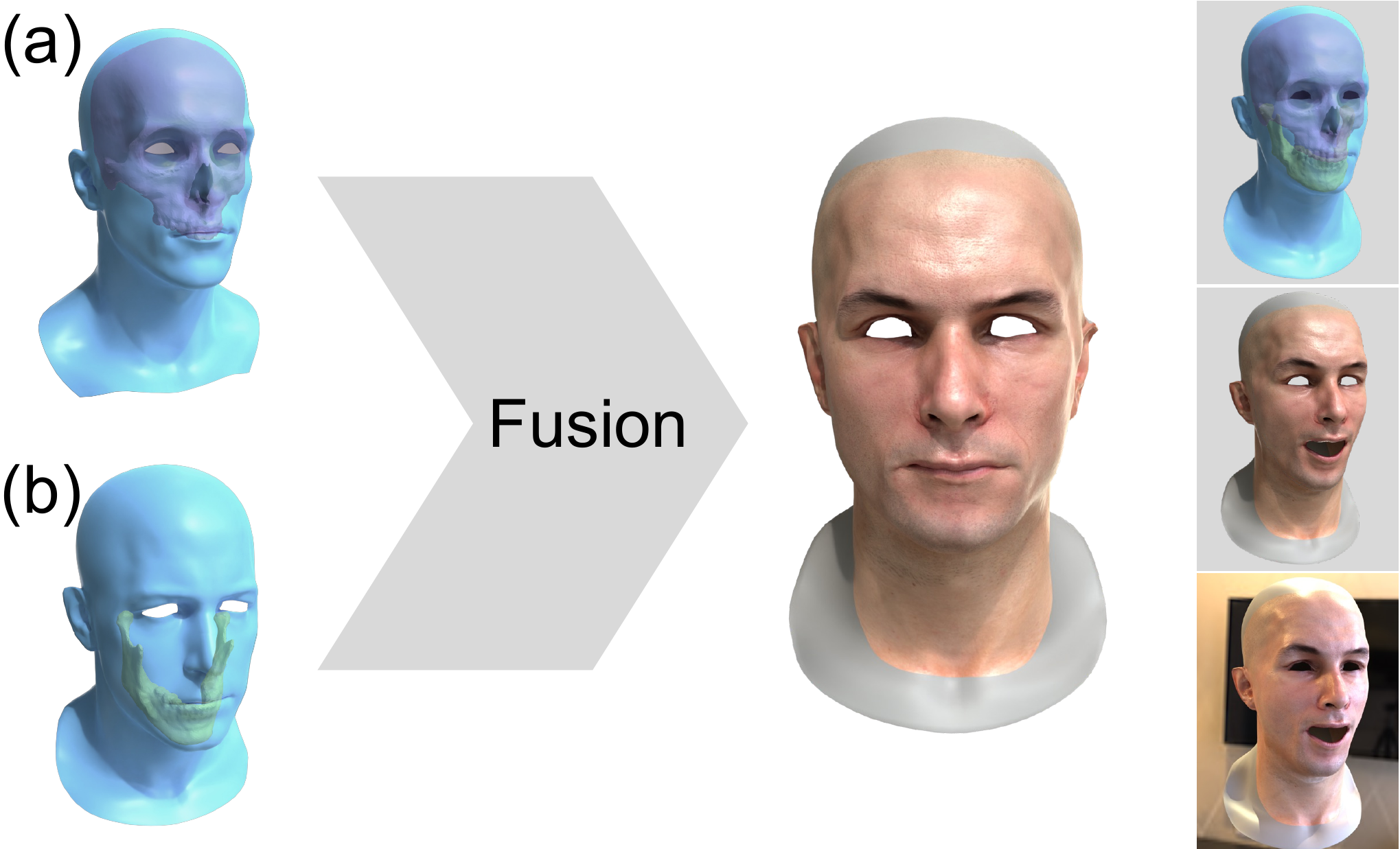}
\caption{Example of character fusion result. The fused face is generated by blending mandible of Ben Affleck (b) into the maxilla of Christian Bale (a) using SCULPTOR. }
\label{fig:fusion}
\end{figure}
%%%%%%%%%%%%%%%%%%%%%%%%%%%%%%%%%%%%%%%%%%%%%%%%%%%%%%%%%%%%%
\paragraph{Qualitative Results.}
%%%%%%%%%%%%%%%%%%%%%%%%%%%%%%%%%%%%%%%%%%%%%%%%%%%%%%%%%%%%%

% Figure \ref{fig:diveristy} 展示了一男一女的identity在diversity component的变化下对于脸局部的变化. 每个人的三行local 放大图分别对应了不同component对脸局部的变化. 第一行从左至右下颌变宽的过程, 第二行展示了下颌从左至右逐渐上下拉伸变长, 第三行反应了下颌逐渐包住上颌(mandibular protrusion的过程). 相比于shape的全局变化, 我们的diversity component能控制脸型的局部变化 从而产生characteristic 的变化

Fig. ~\ref{fig:Gallery2} displays the skeleton-driven characteristic face edition performance on one female and one male actor’s face using the trait space in SCULPTOR. Following the skull curvature process described in Section ~\ref{sec:rendering}. Each row of the partial enlargement displays the characteristic facial variations according to a representative trait component: the first row of jaw width, the second row of jaw length and the third row with relative position between jaw and maxilla. The novel proposed trait space in SCULPTOR extends the parametric model function to control the local skeletal changes and produce accordingly facial characteristic variations.

% 在Figure \ref{fig:texture} 中,我们对两个不同脸型的5个random texture 分别对 左侧的脸型颧骨更高, 下颌更长的男性, 与右侧的脸型较小, 下颌偏圆的女性进行展示. 我们的texture生成能够达到人脸various的效果.

Fig. ~\ref{fig:Gallery3} shows a group of randomly generated facial appearance variations of SCULPTOR. The right three columns show appearance changes on a male face with a relatively higher cheekbone and longer mandible. While the left three columns present a female face with a narrow face contour and round, smooth jaw. As can be seen, our texture space provides realistic color space for face appearance generation.

% \begin{figure*}[t]
%     \centering
%     \includegraphics[width=0.9\linewidth]{NavieFig/Fig12.png}
%     \caption{Gallery1}
%     \label{fig:Gallery1}
%     \end{figure*}

%%%%%%%%%%%%%%%%%%%%%%%%%%%%%%%%%%%%%%%%%%%%%%%%%%%%%%%%%%%%%
\paragraph{Ablation.} 
\label{sec:skull_infer}
%%%%%%%%%%%%%%%%%%%%%%%%%%%%%%%%%%%%%%%%%%%%%%%%%%%%%%%%%%%%%
% to application
% CT face 2 skull
% 相比flame, 我利用术前术后的CT scan将diversity 从shape space分离出来. 为了证明分离出diversity的优势, 我们设计一个diversity的ablation实验. 我们训练的两个SKAVENGER模型, 一个没有分离diversity的SKAVERGER模型是treat72对术前术后的training数据为不同identity, 组成的144个identities, 通过消除pose与expression的影响训练得到的mean template与144个shape  components, denote as SKAVENGER144. 另一个是使用72个training术后数据作为identities, 消除pose与expression的影响训练得到的mean template与72个shape  components, plus 利用72对术前术后的差分得到72个diversity components, denoted as SKAVENGER 72D72. 我们将两个模型都没见过的12对术前术后CT scan数据作为test, 对其展开face reconstruction error的计算, 两个模型各自使用一共144个components对test 数据拟合. 我们以Mean Vertex Squared Error以毫米为单位作为衡量指标. 在表格Table~\ref{tab:skull_inference} 中没有分离diversity space的SKAVENGER144 在术前与术后的数据上的reconstruction error分别是2.01 mm与2.04 mm. 而分离了diversity space的SKAVENGER72D72在两个test集合上都是1.77毫米的误差, 比SKAVENGER144. 并且SKAVENGER72D72比SKAVENGER144,证实了分离出diversity space使得重建效果更好

The novel proposed face trait space is built as an additional local variation based on the parametric shape space. We conduct an ablation study to clarify the effect of using the trait space. We trained a simplified SCULPTOR model that cancels the trait components and uses both pre- and post-surgery CT data from 72 identities for training the shape parameter space. The simplified model is denoted as SCULPTOR-SIMPLE and compared with the standard SCULPTOR in Tab. ~\ref{tab:skull_inference}. \revise{SCULPTOR-SIMPLE uses 144 shape components while SCULPTOR uses 72 shape plus 72 trait components}. We use another 12 pairs pre- and post-surgery CT data to test the two models' skull fitting performance. Mean Vertex Squared Error (MVSE) is used for evaluating the skull fitting results. The averaging MVSE for pre- and post-surgery data from SCULPTOR are both lower than that from the SCULPTOR-SIMPLE model, which demonstrates the effectiveness of adding the trait space in SCULPTOR in elevating the expressiveness of the skeletal structure \revise{and capturing nuanced shape differences}. 

% To evaluate the anatomical correctness of the correlation between face and skull, we evaluate the reconstruction of skull by minimizing the face reconstruction error with $E_{\beta}$ and $E_{\gamma}$ since skull and face shape are trained together, implying they share using $\beta$ and $\gamma$. The final optimization function is $E_{face} + \lambda_b E_{\beta} + \lambda_g E_{\gamma}$, we set $\lambda_b$ and $\lambda_g$ to 0.08. The experiment is performed on the 12 pairs of test pre-operation CT scans dataset and we compare the models using 100 shape components, using 50 shape components and 50 diversity components. We can notice from Table~\ref{tab:skull_inference} in the skull face correlation, with the same number of components using 100 shape components and 50 shape and diversity components respectively, the model of SKAVENGER50D50 performs better than SKAVENGER100D0 on mandible, maxilla and face vertex error. This implies the diversity components could perform well in high diversity data. With the number of shape components increasing to 100, the SKAVENGER100D50 could achieve a smaller overall vertex error, which implies the facial and skull correlated well.

\begin{figure}[t]
\centering
\includegraphics[width=\linewidth]{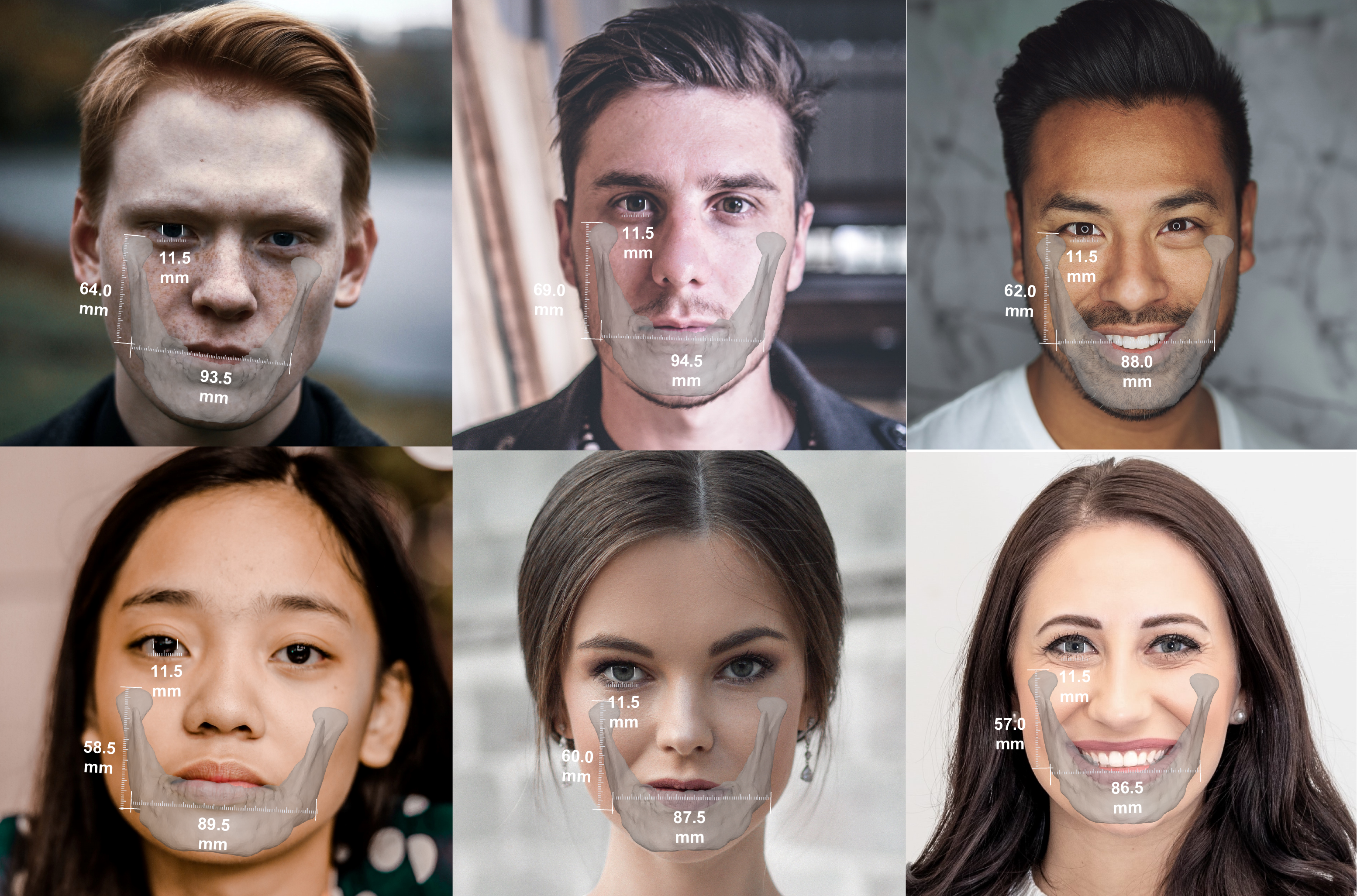}
\caption{\revise{Qualitative result of skull inference from RGB images for actors with a variety of face shapes. Images come from A. Vityukova, T. Barlin, J. Gonzalez,  P. J. Manlapig, T. Bellis and Calyton (Unsplash).}}
\label{fig:RGB2skull}
\end{figure}

%%%%%%%%%%%%%%%%%%%%%%%%%%%%%%%%%%%%%%%%%%%%%%%%%%%%%%%%%%%%%
\paragraph{Comparison with other models.}
% \paragraph{Compare with with 3DMM and DECA on various tasks}
%%%%%%%%%%%%%%%%%%%%%%%%%%%%%%%%%%%%%%%%%%%%%%%%%%%%%%%%%%%%%
% 我们用12个未见过的ct pre, post operation的人脸和未见过的200个来自FaceScape的neutral作为test数据, 来评估我们模型. 我们将我们的模型与flame相比较, FLAME使用了300个shape components与100个expression components, SCULPTOR-1使用了300个shape components与53个expression components,SCULPTOR-2使用了228个shape components, 72个diversity components与53个expression components.

%实验结果如 Figure~\ref{tab:QuantiFaceRecon}所示, 在pre, post 的test集合上 在相同数目的shape components下, SCULPTOR-1的error略微比FLAME低,而使用了72个diversity components 代替72个shape components在both 在pre, post 的test集合上有明显下降. 在FaceScape数据集上,SCULPTOR-1, SCULPTOR-2 perform比FaceScape好, (原因是FaceScape的shape variance比较小,不是实验做错了).
% 在Figure~\ref{fig:qualitative_recon} 中展示了reconstruction的qualitative 的结果.

In the face reconstruction task, we experiment on 12 pairs of CT test data, 200 neutral face mesh test data from FaceScape \cite{yang2020facescape} \revise{and 22 neutral faces from 3DRFE~\cite{3drfe}}. We evaluate SCULPTOR-1 model with 300 shape components and 53 expression components, SCULPTOR-2 model with 228 shape components, 53 expression components and 72 trait components. \revise{SCULPTOR shape components are built from the PCA components on the neutral faces from the post-operation part in LUCY and FaceScape \cite{yang2020facescape} datasets. } To compare with the FLAME model~\cite{flame} with 300 shape components and 100 expression components. 
Table~\ref{tab:QuantiFaceRecon} indicates the quantitative result of the average reconstruction error on each group of test data, reported using facial area RMSE in millimeters (mm) \revise{on pre-, post-surgery and FaceScape dataset. And the result on 3DRFE~\cite{3drfe} dataset (last column) is demonstrated by chamfer distance in millimeters (mm)}. Using the same number and type of principle components, SCULPTOR-1 produces lower errors in both skull and facial mesh fitting tasks than those from FLAME model. While in SCULPTOR-2 model, we replace \revise{the last} 72 shape components with 72 trait components from SCULPTOR-1 model, and SCULPTOR-2 achieves better performance on surgery scans, but the fitting result on FaceScape testset is not as good as that using SCULPTOR-1. This could be caused by the limited variation in face shapes in the FaceScape data. We could visualize the  facial mean squared per-vertex error in Fig~\ref{fig:qualitative_recon}. \revise{On the unseen 3DRFE data, SCULPTOR-2 performs nearly the same as FLAME on chamfer distance.}
% \revise{On the unseen 3DRFE data, SCULPTOR-1,2 perform better than FLAME on chamfer distance, which indicates the effectiveness of using trait component.}

% The high vertex error in Pre-operation scans implies a higher diversity compared to the Post-operation scans.  (这句不要了吧?)
% \emph{SCULPTOR 100 D50} is able to perform better than ICT and FLAME on CT scans with the smallest reconstruction error on both Pre-operation and Post-operation test data. In \todo{[FaceScape]}, our model manage to perform better than  FLAME . Figure~\ref{fig:Facegeneration} illustrates the reconstruction of each model. In comparison with \todo{FLAME}, our model manages to reconstruct the pre-operation and post-operation scans and FaceScape better.

\begin{figure*}[t]
\centering
\includegraphics[width=1\linewidth]{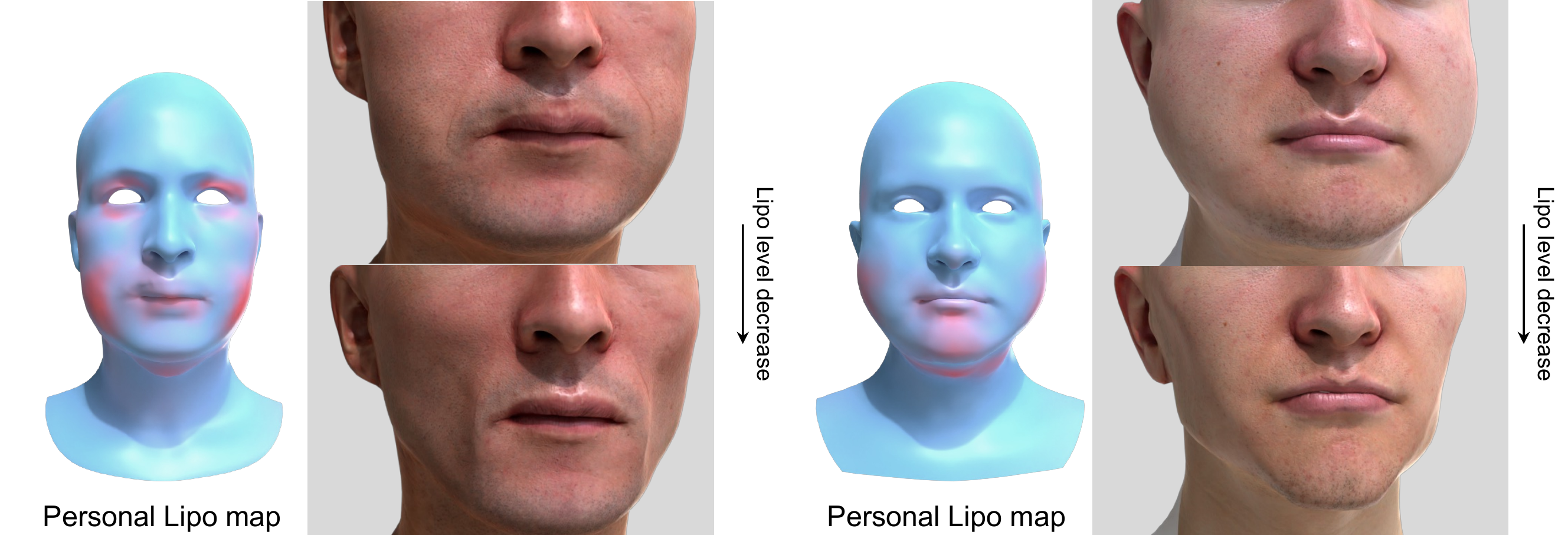}
\caption{Person-specific lipo level change effect. We show facial deformation of two subjects under decreasing lipo levels. The blue models are the personal lipo maps associated with body weight accumulation.}
\label{fig:BMI}
\end{figure*}

% (Dick Cheney)
\begin{figure}[t]
\centering
\includegraphics[width=0.9\linewidth]{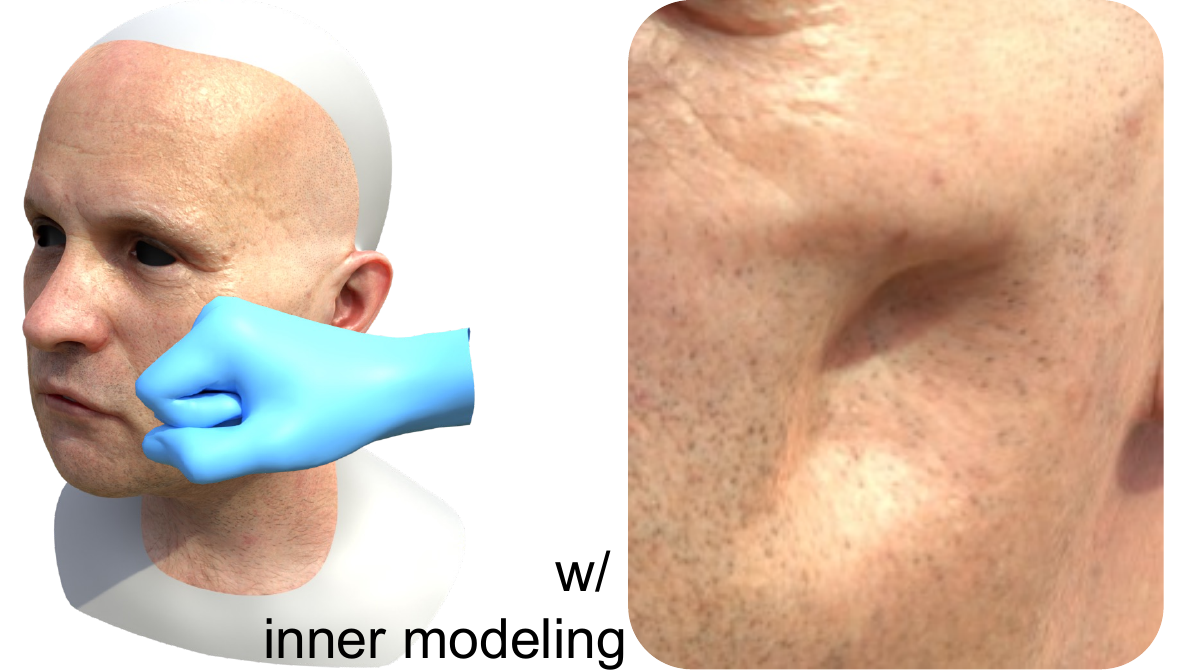}
% \caption{Face simulation. (a)(b) Face hit by a ball with/without skull collision. (c)(d) Face jiggle while shading head with/without skull collision.}
% 只有打脸？
\caption{
    Facial animation using SCULPTOR. A fist punch to the cheekbone results in realistic skin deformation which follows the shape of the maxilla. 
    % artifacts without inner skull modeling(left). More realistic deformations are obtained with our inner model (right).
}
\label{fig:PunchFace}
\end{figure}

%%%%%%%%%%%%%%%%%%%%%%%%%%%%%%%%%%%%%%%%%%%%%%%%%%%%%%%%%%%%%
\subsection{Archaeological Skeletal Facial Completion}
%%%%%%%%%%%%%%%%%%%%%%%%%%%%%%%%%%%%%%%%%%%%%%%%%%%%%%%%%%%%%

%%%%%%%%%%%%% May 18, 23:00 version%%%%
% Figure~\ref{fig:ava} 展示了SCULPTOR对Ava头骨重建以及脸部diversity的渲染展示. Ava 上颌骨与artist [Hew Morrison]重建的脸分别在Figure~\ref{fig:ava}(a)的左下角展示与右下角,  我们的参数化模型有能力通过energy function来恢复Ava她的下颌骨, (a)的上侧是我们使用SCULPTOR100的100个shape components来优化Ava的上颌骨, 得到的Ava的脸型与其对应的上颌骨, (b)是Ava的下巴的正视图与侧视图, 侧视图中叠加了regress的下颌骨.

% 然后我们用diversity basis ,使用不同的init gamma_0 来拓展Ava下巴位置处的的characteristic. 我们分别使用下巴变宽, which可以从(c)图中下侧看出, 这使得(c)图Ava的下半张脸更加宽. 在(d)图中, 我们利用初始化变窄的下颌骨的gamma_0使得最后生成的脸比(b), (c)中的窄, 脸型变小.
% 这些qualitative的图表示了SCULPTOR模型的拓展characteristic的能力. 

Fig.~\ref{fig:ava} presents the face generation results recovered from the maxilla of Ava. The original Ava maxilla and face constructed by forensic artists are shown in Fig.~\ref{fig:ava}(a). Our parametric model is able to fit with the maxilla and recover the mandible and face appearance for Ava by optimizing $\beta$. 

Fig.~\ref{fig:ava}(a) depicts our reconstructed Ava face with respect to the optimization on maxilla using 100 components in SCULPTOR shape space without using trait components. Then by varying the trait parameter $\gamma$, we generate a variety of face appearance and mandible shapes that make each Ava looks different. 
The visualization results are shown in Fig.~\ref{fig:ava}(b), (c) and (d). Our reconstructed Ava's face in (b) has a round and short chin, the chin in (c) is wide and the face is flat, while the face in (d) is narrow and long. The two figures in the same column show the same face generation rendered from different views. Our model provides high characteristic variation space to generate characteristic faces.
\revise{Moreover, we conducted the skull completion task on 12 pre- and post-surgery test data. The quantitative evaluation results are shown in Tab.~\ref{tab:MissingJaw}. Given the face and part of the skull, SCULPTOR is performed to infer the missing mandible or maxilla. Hausdorff distance between the inference and the missing structure is reported in millimeters in Tab.~\ref{tab:MissingJaw}. }

% Figure~\ref{fig:ava} 展示了SCULPTOR对Ava头骨重建以及脸部diversity的渲染展示. Ava 上颌骨与artist [Hew Morrison]重建的脸分别在Figure~\ref{fig:ava}(a)的左下角展示与右下角,  我们的参数化模型有能力通过energy function来恢复Ava她的下颌骨, (a)的上侧是我们使用SCULPTOR100的100个shape components来优化Ava的上颌骨, 得到的Ava的脸型与其对应的上颌骨, (b)是Ava的下巴的正视图与侧视图, 侧视图中叠加了regress的下颌骨.

% 然后我们用diversity basis ,使用不同的init gamma_0 来拓展Ava下巴位置处的的characteristic. 我们分别使用下巴变宽, which可以从(c)图中下侧看出, 这使得(c)图Ava的下半张脸更加宽. 在(d)图中, 我们利用初始化变窄的下颌骨的gamma_0使得最后生成的脸比(b), (c)中的窄, 脸型变小.
% 这些qualitative的图表示了SCULPTOR模型的拓展characteristic的能力. 在实验中, 我们设置lambda_gamma_0为0.4. 

%Fig.~\ref{fig:ava} shows the face generation results recovered from maxilla of Ava. The original Ava maxilla is shown in Figure~\ref{fig:ava}(a). Our parametric model is able to fitting the maxilla and recover the mandible by optimizing $\beta$. 

%Fig.~\ref{fig:ava}(a) depicts the reconstruction of the Ava face with respect to the optimization on maxilla without using diversity components. Then by varying $\gamma$, the mandible is modified to a variety of shapes. We display different face appearance and mandible generations and the  in Fig.~\ref{fig:ava}(c). With different $\gamma$ initialization as prior to make Ava looks differently, the results visualization are shown in   Fig.~\ref{fig:ava}(b)(c). The two figures in the same column show the same generated person rendered from different views. Our model has high diversity to generate a certain number of faces with respect to the fixed maxilla.

%%%%%%%%%%%%%%%%%%%%%%%%%%%%%%%%%%%%%%%%%%%%%%%%%%%%%%%%%%%%%
\subsection{Character Fusion}
%%%%%%%%%%%%%%%%%%%%%%%%%%%%%%%%%%%%%%%%%%%%%%%%%%%%%%%%%%%%%
% Figure~\ref{fig:fusion}展示了character fusion的结果,我们将蝙蝠侠中的Christian Bale(左上图)与Ben Affleck(左下图)作角色融合, 我们将Ben Affleck的下颌融入给Christian Bale的上颌, 得到fusion的结果. Our model 拟合出的Christian Bale的头骨有较短的上颌, 有又尖又长的下颌, 而Ben Affleck有较短偏圆的下颌, 有较长的上颌. 通过优化 pose, shape, gamma参数得到fusion的结果.  可以看出fusion后生成的的人的拥有与Christian Bale相似的瘦的上脸, 原本Christian Bale尖下巴也因换成了Ben Affleck有较短偏圆的下颌,变的圆. 并且我们让生成的人做张嘴与上挑眉毛的动作和表情并在环境光照下渲染, 得到一个栩栩如生的人.

%We carry out the character fusion task with two actors with obvious different shapes in mandible and maxilla. The results are depicted in Figure~\ref{fig:fusion}.  Finally, we repose the and add certain expression to new face, which looks more vividly.
As shown in Fig.~\ref{fig:fusion}, we carry out the character fusion task with Christian Bale (top left) and Ben Affleck (bottom left) in Batman. Maxilla and mandible shapes of the two actors are obviously different. 
Our model fits Christian’s skull with a short maxilla and a pointed and long mandible. While Ben has a short, rounded mandible and a longer maxilla. 
The fused face is generated by blending Ben's jaw into Christian's maxilla and optimizing the pose, shape, and trait parameters in SCULPTOR. 
The generated face has a slim upper face that is similar to Christian, and the originally pointed chin is replaced by the short round chin of Ben. 
Then we add pose and expression of opening mouth and raising eyebrows to the fused face, and render them under ambient lighting to get a realistic person.

% \todo{======placeholder==========}

%placeholder placeholder placeholder placeholder placeholder placeholder placeholder placeholder placeholder placeholder placeholder placeholder placeholder placeholder placeholder placeholder placeholder placeholder placeholder placeholder placeholder placeholder placeholder placeholder placeholder placeholder placeholder placeholder placeholder placeholder placeholder placeholder placeholder placeholder placeholder placeholder placeholder placeholder placeholder placeholder placeholder placeholder placeholder placeholder placeholder placeholder placeholder placeholder placeholder placeholder placeholder placeholder placeholder placeholder placeholder placeholder placeholder placeholder placeholder placeholder placeholder placeholder placeholder placeholder placeholder placeholder placeholder placeholder placeholder placeholder placeholder placeholder placeholder placeholder placeholder placeholder placeholder placeholder 

% \todo{============placeholder==========}

%%%%%%%%%%%%%%%%%%%%%%%%%%%%%%%%%%%%%%%%%%%%%%%%%%%%%%%%%%%%%
\subsection{Skull Inference from Image}
%%%%%%%%%%%%%%%%%%%%%%%%%%%%%%%%%%%%%%%%%%%%%%%%%%%%%%%%%%%%%
% 我们在Figure~\ref{fig:RGB2skull}展示了使用SCULPTOR的construct脸的结果图. Figure~\ref{fig:RGB2skull} (a)(b)(c)(d)展示了从RGB图像inference下颌骨位置. 这些图像来自于in the wild. 我们adopt在我们拓扑下训练的DECA网络[DECA]输出的脸mesh, 我们采用SCULPTOR100D50, 优化face的global rotation,下颌骨的pose,shape and diversity through SGD optimizer. Finally, 我们对posed重建出的下颌骨通过DECA计算得出的相机外参投影回图片

% Personalized facial features help people recognize others in 0.3 seconds. 
The most important and variable feature for facial contouring is the shape and position of the mandible. People with a square face tend to have a wider mandible, while people with a long mandible tend to have a sharper chin.
% We use SCULPTOR with 500 shape and 50 trait components to $\beta$ $\gamma$, $\theta$ and $\phi$ to regress the face geometry and obtain the internal skeletal structures of each face and we project the mandible back to the images. 
We present the qualitative skull inference results using SCULPTOR with 500 shape and 50 trait components and projecting the internal mandible of each face back to the images in Fig.~\ref{fig:RGB2skull} to analyze each human's characteristic face. %It shows examples of mandible inference from RGB images to find out personalized facial feature.  
In order to accurately compare the facial differences in digital photographs, we adopt "Iris Ruler" as the scale to measure the distance between face and camera when capturing the photo ~\cite{DRIESSEN2011579}. Each person's iris is 11.5mm wide, and there is almost no difference in race, sex and age. It can well compare the width and length of the mandible.
We measure two items: Mandible Width (MW), which means the distance between the left and right mandibular angle, and Ramus Height (RH), which means the distance between mandibular angle and condylar. \revise{We demonstrate the measurement results in the Fig.~\ref{fig:RGB2skull}. Benefiting from the detailed 3D Face recovery using DECA, SCULPTOR is able to reconstruct mandible position with high accuracy.}
\subsection{Face Generation with Lipo Level Change Effect}
%%%%%%%%%%%%%%%%%%%%%%%%%%%%%%%%%%%%%%%%%%%%%%%%%%%%%%%%%%%%%

%
%%%%
% 根据section 7.6的算法, 我们可以实现非常自然的BMI变化下的头部形变, 结果展示在图11中, 每一行代表一个subject, 我们分别在他们的neutral脸上叠加了他们的头骨的渲染。最左边一列是两个subject的personal fat map, 值越大表示对应的脸部区域长胖时效果最明显。由图可见, 随着BMI递减, 意味着subject从胖变瘦, fatmap中数值比较大的部分移动更多。脸颊部分尤为明显, 而头顶和脖子的变化就不太明显。这也是符合人体解剖学规律的, 脸颊部分更容易囤积脂肪\cite{}. 
% 在这个任务中, 虽然只有外表面的脸发生了形变, 但是如果只是简单的移动外表面的顶点, 很容易产生不自然的效果, 很可能会跟内部的骨骼穿模。因为我们的模型有内部的骨骼, 所以可以计算出anatomically correct fat map, 并且在对脸部的顶点做形变的时候实时与骨骼做碰撞检测, 保证脸部永远在骨头外面, 从而实现更加自然的减重效果, 以及对极瘦效果的模拟。
%
%%%%
%
Based on Section \ref{sec:bmi}, we are able to achieve anatomically \revise{consistent} face deformation results when lipo level changes. As shown in Fig. \ref{fig:BMI}, the left and right parts represent two individual subjects. 
The leftmost column of each part shows the personal lipo map, and the values specify which areas of the head are more prone to body weight accumulation.  
As can be observed, as the subject's lipo level drops, the vertices with higher values on the lipo map shift closer to the skull. The impact is most noticeable on the cheeks, whereas the neck and top of the head are less noticeable. This is congruent with human anatomy, as the cheeks are more prone to weight accumulation than the top of the head and neck\cite{swift2021facial}.
Despite the fact that this deformation only changes the surface geometry, simply altering skin vertices would result in unnatural deformation and visual artifacts caused by skin and skull collision. Since our model includes an inside skull, we can generate anatomically consistent lipo maps and use collision detection when deforming face vertices to ensure that the face is always outside the skull, thus achieving a more natural weight loss effect and the slimmest possible head shape.
%%%%%%%%%%%%%%%%%%%%%%%%%%%%%%%%%%%%%%%%%%%%%%%%%%%%%%%%%%%%%
\subsection{Facial Animations }
%%%%%%%%%%%%%%%%%%%%%%%%%%%%%%%%%%%%%%%%%%%%%%%%%%%%%%%%%%%%%

{
Fig.~\ref{fig:PunchFace} shows simulated facial deformation under a fist punch to the cheekbone.
We apply physical simulation on SCULPTOR. 
Since SCULPTOR models the inner skull; it acts as a constraint that prevents skin from penetrating the skull, thus reproducing photorealistic skin-skull collision under motion. 
% and the outer surface only. 
%
% When external forces induce large skin deformation,
Meanwhile conventional surface models without inner skeletal structures will lead to unrealistic skin sunken artifacts.
Please see the comparison sequence in the accompanying video. 
}

%% file: sections/conclusion.tex
\section{Conclusion}

3D human face modeling has attracted increasing attention over the last few years. Most existing approaches overwhelmingly focused on modeling the exterior shapes, textures and skin properties of faces to generate faithful human faces. However, these methods ignored the anatomic facial bone structures in the model generation process. In this paper, we proposed LUCY, a comprehensive shape-skeleton correlated face dataset from pre-and post-surgery CT imaging and 3D scans. To better explore the inherent correlation between inner skeletal structures and appearance, we developed SCULPTOR, a novel skeleton consistent parametric facial generator that jointly models the skull, face geometry and face appearance under a unified data-driven framework. SCULPTOR preserves both anatomic correctness and visual realism in facial generation tasks compared with existing methods. The robustness and effectiveness of SCULPTOR have been clearly shown in various applications, e.g. Archaeological Skeletal Facial Completion, Character Fusion, Skull Inference from Image, Face Generation with Lipo Level Change Effect and Physical Simulation. In particular, the skeleton-consistent nature of our model design can enrich currently scarce 3D face data with physical correctness. 
% To generate faithful and realistic 3D human face, most existing approaches
% overwhelmingly emphasized on modeling the exterior shapes, textures and skin properties of faces. However, ignoring the anatomic facial bone structures in the model generation process. In this work, we propose LUCY, the first comprehensive shape-skeleton correlated face dataset from pre-and post-surgery CT imaging and 3D scans. To better explore the inherent correlation between inner skeletal structures and appearance, we presents SCULPTOR, a novel skeleton consistent parametric facial generator that jointly models the skull, face geometry and face appearance under a unified data-driven framework. SCULPTOR preserves both anatomic correctness and visual realism in facial generation tasks compared with existing methods. We demonstrate the model's robustness and effectiveness in various applications, e.g. Archaeological Skeletal Facial Completion, Character Fusion, Skull Inference from Image, Face Generation with Lipo Level Change Effect and Physical Simulation. In particular, the skeleton consistent nature of our model design can enrich currently scare 3D face data with physical correctness.

As we analyzed before, SCULPTOR has several strong points on jointly modeling the correlation of skull and face as well as characteristic variance. We also want to highlight the potential drawbacks of this approach. As the data in LUCY only comes from plastic surgery, which often changes the mandible, zygomatic and maxillary alveolar bone locally, the diversity of its distribution may be limited. The deformation of the trait component depends entirely on whether a certain part of the bone will be surgically designed during the operation. Therefore, the 
trait component of our model is limited to these corresponding parts. In future work, we plan to enrich our LUCY dataset with the registration of existing off-the-shelf datasets, and further enhance the characteristic variance of our model.
Besides, our model uses Linear Blend Skinning and expression blend shape without muscle modeling. In the future, we plan to collect the facial expression data with dynamic MRI, as MRI can quickly scan the face to obtain muscles and soft tissue information without radiation. Jointly modeling the skull, muscles, face geometry and face appearance to construct a 3D human face will be an interesting area and need to be well explored.
% Although our SCULPTOR can jointly models the correlation of skull and face as well as characteristic variance, it also have some limitation due to the diversity of surgical data is limited. The deformation of the diversity component depends entirely on whether a certain part of the bone will be surgically designed during the operation. As LUCY comes from plastic surgery, which often changes the mandible, zygomatic and Maxillary alveolar bone locally. Therefore, the diversity component of our model is limited to these corresponding parts. In future work, we plan to enrich our LUCY dataset with the registration of existing off-the-shelf datasets.